\documentclass[10pt,twocolumn,letterpaper]{article}

\usepackage{amsmath,amsfonts,bm}

\def\eqref#1{equation~\ref{#1}}

\def\1{\bm{1}}

\def\vf{{\bm{f}}}

\def\vw{{\bm{w}}}
\def\vx{{\bm{x}}}
\def\vy{{\bm{y}}}

\DeclareMathAlphabet{\mathsfit}{\encodingdefault}{\sfdefault}{m}{sl}
\SetMathAlphabet{\mathsfit}{bold}{\encodingdefault}{\sfdefault}{bx}{n}

\usepackage{cvpr}      

\usepackage[accsupp]{axessibility}
\usepackage{graphicx,adjustbox}
\usepackage{url}
\usepackage{amssymb}
\usepackage[linesnumbered,ruled,vlined]{algorithm2e}
\SetAlFnt{\small}
\SetKwComment{Comment}{/* }{ */}
\SetKwInput{kwInput}{Input}
\SetKwInput{kwOutput}{Output}
\RestyleAlgo{ruled}
\usepackage{amsmath}
\usepackage{array}
\usepackage{multirow}
\usepackage{booktabs}
\usepackage[table]{xcolor}
\usepackage[normalem]{ulem}
\useunder{\uline}{\ul}{}
\usepackage{float, wrapfig}
\usepackage{resizegather}
\usepackage{enumitem}
\def\code#1{\texttt{#1}}
\usepackage{diagbox}

\usepackage[font=small,skip=0pt]{caption}
\usepackage{amssymb}

\usepackage[pagebackref,breaklinks,colorlinks]{hyperref}
\usepackage[capitalize]{cleveref}
\crefname{section}{Sec.}{Secs.}
\Crefname{section}{Section}{Sections}
\Crefname{table}{Table}{Tables}
\crefname{table}{Tab.}{Tabs.}

\def\cvprPaperID{14016} 
\def\confName{CVPR}
\def\confYear{2024}
\begin{document}

\title{Large-Scale Data-Free Knowledge Distillation for ImageNet via Multi-Resolution Data Generation}

\author{Minh-Tuan Tran$^1$, Trung Le$^1$, Xuan-May Le$^2$, Jianfei Cai$^1$, Mehrtash Harandi$^1$, Dinh Phung$^{1,3}$\\\
$^1$Monash University,  $^2$University of Melbourne, $^3$VinAI Research\\
{\tt\small \{tuan.tran7,trunglm,mehrtash.harandi,jianfei.cai,dinh.phung\}@monash.edu} \\ \tt\small xuanmay.le@student.unimelb.edu.au}
\maketitle

\begin{abstract}

Data-Free Knowledge Distillation (DFKD) is an advanced technique that enables knowledge transfer from a teacher model to a student model without relying on original training data. While DFKD methods have achieved success on smaller datasets like CIFAR10 and CIFAR100, they encounter challenges on larger, high-resolution datasets such as ImageNet. A primary issue with previous approaches is their generation of synthetic images at high resolutions (e.g., $224 \times 224$) without leveraging information from real images, often resulting in noisy images that lack essential class-specific features in large datasets. Additionally, the computational cost of generating the extensive data needed for effective knowledge transfer can be prohibitive. In this paper, we introduce \underline{MU}lti-Re\underline{S}olution Data-Fre\underline{E} (MUSE) to address these limitations. MUSE generates images at lower resolutions while using Class Activation Maps (CAMs) to ensure that the generated images retain critical, class-specific features. To further enhance model diversity, we propose multi-resolution generation and embedding diversity techniques that strengthen latent space representations, leading to significant performance improvements. Experimental results demonstrate that MUSE achieves state-of-the-art performance across both small- and large-scale datasets, with notable performance gains of up to two digits in nearly all ImageNet and subset experiments. Code is available at \url{https://github.com/tmtuan1307/muse}.

\end{abstract}

\section{Introduction}

Knowledge distillation (KD) \cite{kd,kd1,kd2,kd3} is a technique aimed at training a student model to replicate the capabilities of a pre-trained teacher model. Over the past decade, KD has been applied across various domains, including image recognition \cite{kd3}, speech recognition \cite{skd}, and natural language processing \cite{nlpkd}. Traditional KD methods typically assume that the student model has access to all or part of the teacher's training data. However, in many real-world scenarios, particularly in privacy-sensitive fields like healthcare, accessing the original training data is not feasible due to legal, ethical, or proprietary constraints. In such cases, conventional KD methods become impractical, necessitating alternative approaches that do not rely on direct access to the original data.

To address this challenge, Data-Free Knowledge Distillation (DFKD) \cite{adi,cmi,spshnet,lander,kakr,mad,dfkd_gr} has emerged as a promising solution. DFKD transfers knowledge from a teacher neural network ($\mathcal{T}$) to a student neural network ($\mathcal{S}$) by generating synthetic data instead of using the original training data. This synthetic data enables adversarial training between the generator and the student \cite{zskd,zskt}, where the student aims to match the teacher's predictions on the synthetic data, while the generator's objective is to create samples that maximize the discrepancy between the teacher's and student’s predictions.

\begin{figure}[t]
\begin{center}
\includegraphics[width=\linewidth]{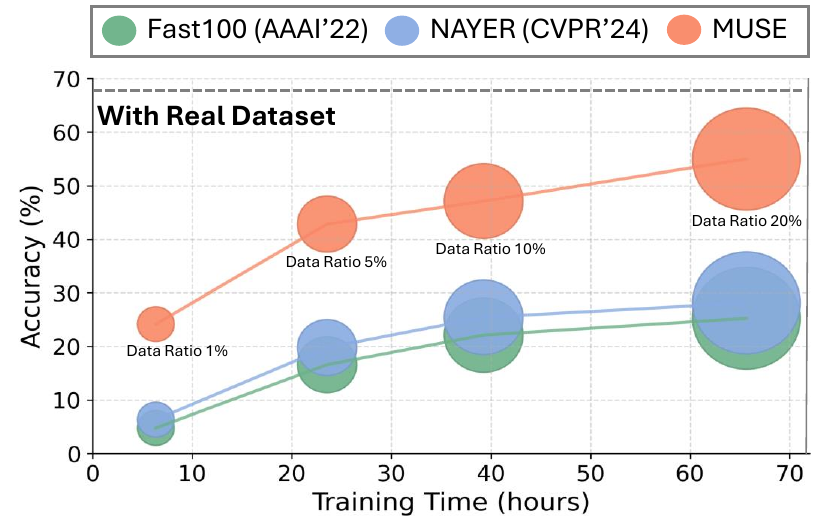}
\end{center}
\caption{Accuracies of our MUSE method and current SOTA Fast \cite{fastdfkd} and NAYER \cite{nayer} on ImageNet1K, all evaluated under approximate training time and same data scale ratios from 1, 5, 10 and 20\% of the original training set (bubble size).}
\label{fig:intro}
\end{figure}

Previous works \cite{adi, cmi, spshnet, lander, kakr, mad, dfkd_gr} typically generate synthetic data at the same resolution as the images used to train the teacher model, a technique that has proven effective on small-scale datasets like CIFAR10 and CIFAR100 \cite{nayer, fastdfkd}. However, these approaches face significant challenges when applied to larger, high-resolution datasets such as ImageNet. A primary issue with previous methods is their generation of synthetic images at high resolutions (e.g., $224 \times 224$) without incorporating information from real images, leading to substantial noise and a lack of nuanced, class-specific features critical for effective knowledge transfer. Additionally, the computational cost of generating the large volumes of synthetic data required for knowledge transfer can be prohibitively high. For instance, previous methods \cite{adi} have demanded over 3,000 GPU hours to train on ImageNet1k, yet have achieved only moderate results. As a result, while DFKD methods perform well on smaller datasets, they encounter substantial limitations when scaled to real-world, large-scale applications.

\begin{figure}[t]
\begin{center}
\includegraphics[width=\linewidth]{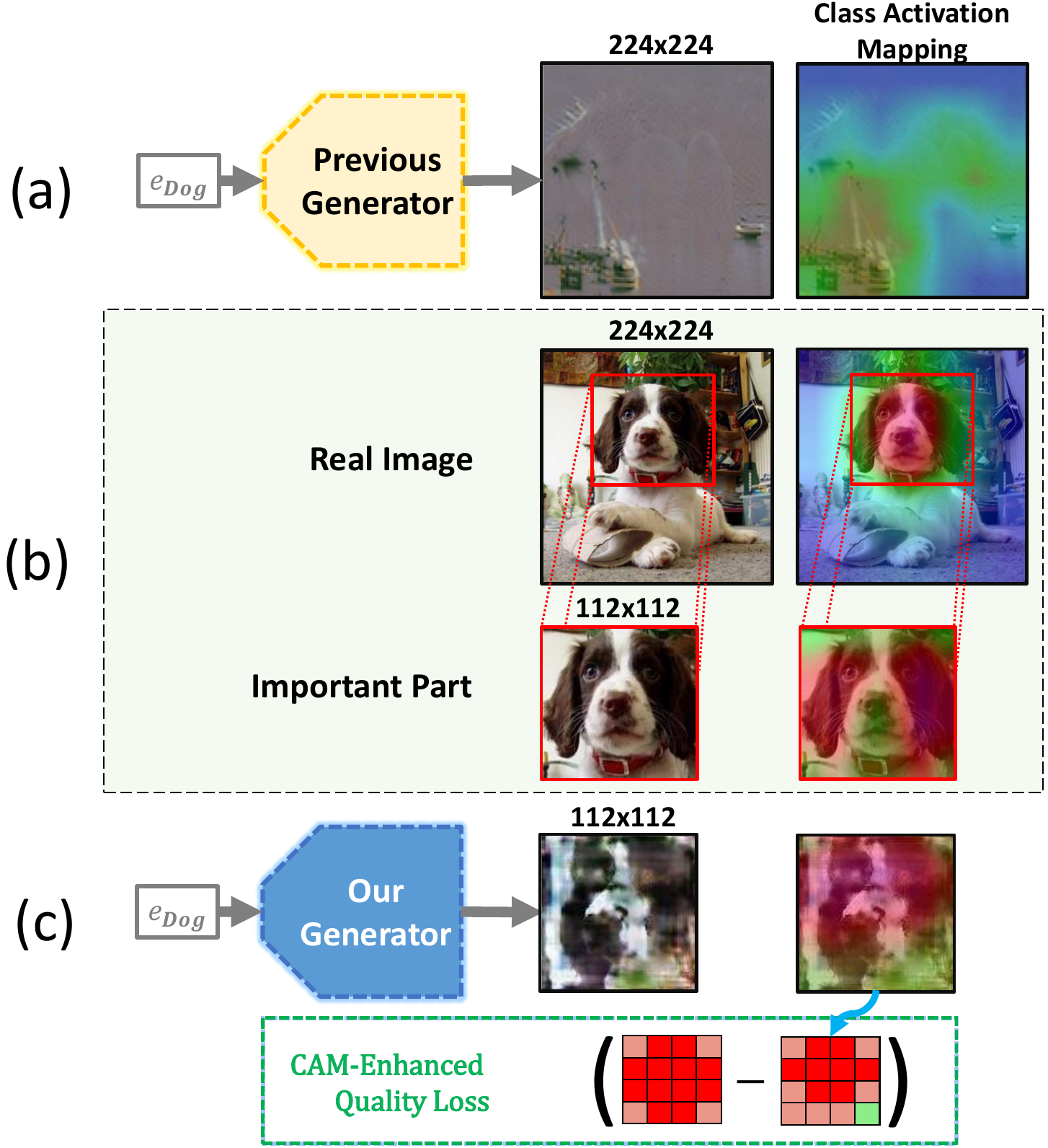}
\end{center}
\caption{(a) The previous model fails to capture class-specific features and contains a lot of noisy pixels. (b) The visualization demonstrates that only a small set of key features is important for classifiers. (c) Our model generates synthetic images at lower resolutions and leverages CAM to generate pixels containing important information. }
\label{fig:diff}
\end{figure}

\begin{figure}[t]
\begin{center}
\includegraphics[width=\linewidth]{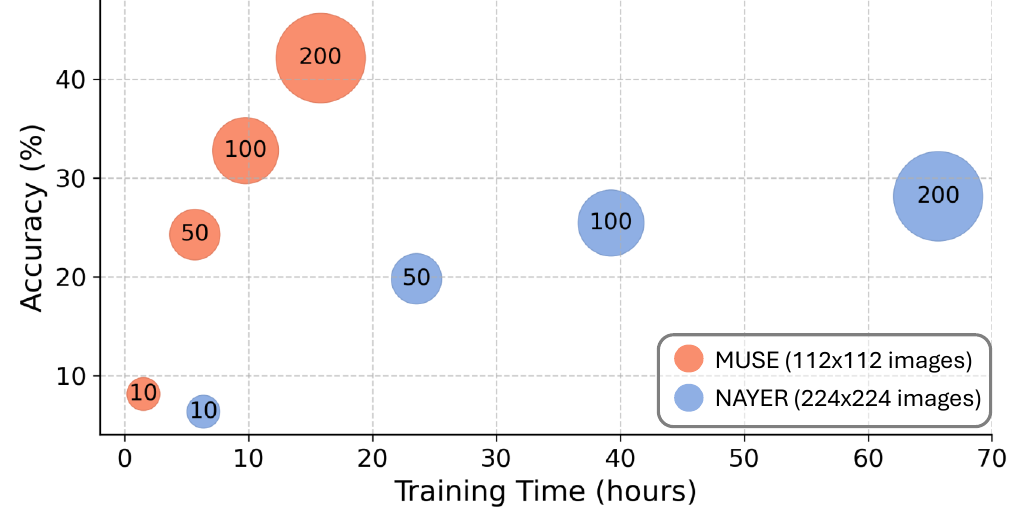}
\end{center}
\caption{Accuracies and training time of using lower-resolution $112\times 112$ images (MUSE) are compared to higher-resolution $224 \times 224$ images (NAYER \cite{nayer}) on ImageNet1K with various images-per-class settings (bubble size). It is clear that using lower-resolution images not only improves performance but also significantly speeds up the training time.}
\label{fig:moti1}
\end{figure}

In this paper, we introduce \textit{\underline{MU}lti-Re\underline{S}olution Data-Fre\underline{E}} Knowledge Distillation (MUSE) to tackle the limitations of traditional DFKD methods. Motivated by the observation that only a small, yet crucial region of real images is essential for classifier training \cite{detr, gradcam, visual1}, MUSE generates synthetic images at lower resolutions and leverages Class Activation Maps (CAMs) \cite{cam} to focus on generating pixels that contain this key information. By concentrating on the most relevant areas, MUSE ensures that the generated images retain critical class-specific features, thereby improving the efficiency of knowledge transfer. Additionally, this approach helps reduce computational costs, enhancing both the scalability and performance of DFKD, especially for large, high-resolution datasets. As shown in Figure \ref{fig:diff}, the previous DFKD model generates $224 \times 224$ resolution images, which are often noisy and provide limited information for training the classifier. In contrast, our method produces lower-resolution images that leverage a CAM-enhanced loss to retain discriminative features. Moreover, as illustrated in Figure \ref{fig:moti1}, our method not only significantly speeds up training time but also achieves improved accuracy across various image-per-class settings, demonstrating its efficiency and effectiveness.

Although using lower-resolution synthetic images improves training efficiency, it may limit the model's ability to capture diverse and detailed feature representations, as lower resolutions constrain the space available for such diversity. To address this limitation, we propose a multi-resolution generation approach, where images are generated at multiple resolutions to capture both coarse and fine-grained features. Additionally, we introduce embedding diversity techniques to preserve distinctiveness within the latent space, ensuring that rich feature representations are maintained even at lower resolutions. These designs allow the model to retain crucial features at varying levels of granularity, leading to improved performance and robustness across a wide range of tasks.

Our major contributions are summarized as follows:
\begin{itemize}[noitemsep,nolistsep]
\item We propose Multi-resolution Data-Free Knowledge Distillation (MUSE), which generates synthetic images at lower resolutions, using Class Activation Maps to focus on critical regions, improving computational efficiency without sacrificing essential class-specific features.
\item To capture both coarse and fine-grained features, we introduce a Multi-Resolution Generation approach, enhancing feature diversity and model performance.
\item We propose Embedding Diversity techniques to preserve distinctiveness in the embedding space, ensuring rich representations even at lower resolutions, thereby improving generalization.
\item MUSE achieves state-of-the-art performance on both small- and large-scale datasets, including CIFAR10, CIFAR100, ImageNet, and its subsets. Our method demonstrates performance gains of up to two digits in nearly all experiments on ImageNet and its subsets.
\end{itemize}

\section{Related Work}
\noindent
\textbf{Data-Free Knowledge Distillation.} DFKD methods \cite{adi, cmi, spshnet, mad, kakr} generate synthetic images to facilitate knowledge transfer from a pre-trained teacher model to a student model. These synthetic data are used to jointly train the generator and the student in an adversarial manner \cite{zskt}. Specifically, the student aims to make predictions that closely align with the teacher’s on the synthetic data, while the generator strives to create samples that match the teacher's confidence while also maximizing the mismatch between the student’s and teacher’s predictions. This adversarial process fosters a rapid exploration of synthetic distributions that are valuable for knowledge transfer between the teacher and the student.

\noindent
\textbf{Data-Free Knowledge Distillation for Imagenet.} Data-free knowledge distillation methods face significant challenges when scaled to larger, high-resolution datasets like ImageNet. For instance, DeepInv \cite{adi} required over 3000 NVIDIA V100 GPU hours to train on ImageNet1k, highlighting the substantial computational demands. Although more recent methods \cite{nayer, fastdfkd} provide faster solutions, they cannot achieve competitive performance when training models from scratch without the pretrained data used by DeepInv. Therefore, there is an urgent need for novel methods that can efficiently and effectively enable data-free transfer on large-scale datasets like ImageNet.

\section{Proposed Method}
\subsection{Preliminaries: DFKD Framework}

Consider a training dataset $D = \{(\vx_i, \vy_i)\}_{i=1}^m$, where each $\vx_i \in \mathbb{R}^{c \times h \times w}$ is an input sample and $\vy_i \in \{1, 2, \dots, K\}$ denotes its label. Each pair $(\vx_i, \vy_i)$ in $D$ serves as a training example with its corresponding label. Let $\mathcal{T} = \mathcal{T}_{\theta_\mathcal{T}}$ represent a pre-trained teacher network on $D$. The objective of DFKD is to train a student network, $\mathcal{S} = \mathcal{S}_{\theta_\mathcal{S}}$, to match the teacher's performance without access to the original dataset $D$.

To achieve this, inspired by \cite{nayer}, given a batch of pseudo-labels $\hat{\vy}$, we first query its text embedding in the label using a pre-trained language model $\mathcal{C}$, i.e., $\vf_{\vy} = \mathcal{C}(\hat{\vy})$. After that, $\vf_{\vy}$ is fed to the noisy layer $\mathcal{Z}$ and a lightweight generator $\mathcal{G}$ to create synthetic images $\hat{\vx}$.
\begin{equation}
    {\hat{\vx}} = {\mathcal{G}}_{l \times l}(\mathcal{Z}(\vf_{\vy})),
\end{equation}
where $\hat{\vx} \in \mathbb{R}^{l \times l}$, with $l$ representing the resolution of the training data (e.g., $224 \times 224$ for ImageNet or $32 \times 32$ for CIFAR10/CIFAR100). Note that we use ${\mathcal{G}}_{l \times l}$ to specify the generator that produces $l \times l$ resolution images. Subsequently, $\hat{\vx}$ is stored in a memory pool $\mathcal{M}$ and used to jointly train both the generator and the student network in an adversarial setup \cite{zskt}. In this setup, the student is trained to approximate the teacher's predictions on synthetic data by minimizing the Kullback-Leibler (KL) divergence loss between $\mathcal{T}(\hat{\vx})$ and $\mathcal{S}(\hat{\vx})$.
\begin{align}
    \hat{\vy}_{\mathcal{S}} &= \mathcal{S}(\hat{x}); \hat{\vy}_{\mathcal{T}} = \mathcal{T}(\hat{x})\; \nonumber \\
    \mathcal{L}_{\mathcal{S}} &= \mathcal{L}_{\text{KL}} = KL(\hat{\vy}_{\mathcal{T}}, \hat{\vy}_{\mathcal{S}}), 
    \label{eq:kl}
\end{align}
while the generator aims to produce samples that not only align with the teacher's confidence but also maximize the discrepancy between the student’s and teacher’s predictions.
\begin{align}
    \mathcal{L}_{\mathcal{G}} &= \alpha_{ce}\mathcal{L}_{ce} + \alpha_{adv}\mathcal{L}_{adv} + \alpha_{bn}\mathcal{L}_{bn} \nonumber \\ 
    &= \alpha_{ce}\mathcal{L}_\text{CE}(\hat{\vy}_{\mathcal{T}}, \hat{\vy}) -\alpha_{adv}KL(\hat{\vy}_{\mathcal{T}},\hat{\vy}_{\mathcal{S}}) + \alpha_{bn}\mathcal{L}_\text{BN}(\mathcal{T}(\hat{\vx}))
\end{align}
In this framework, $\mathcal{L}_{ce}$ represents the Cross-Entropy loss, training the student on images within the teacher's high-confidence regions. In contrast, the negative $\mathcal{L}_{adv}$ term encourages exploration of synthetic distributions, enhancing knowledge transfer from the teacher to the student. Here, the student network acts like a discriminator in GANs, guiding the generator to produce images that the teacher has mastered but the student has yet to learn, thereby focusing the student’s development on areas where it lags behind the teacher. Additionally, we apply batch norm regularization ($\mathcal{L}_{bn}$) \cite{adi, fastdfkd}, a standard DFKD loss, to align the mean and variance at the \texttt{BatchNorm} layer with its running mean and variance. This adversarial setup facilitates the efficient exploration of synthetic distributions for effective knowledge transfer between the teacher and the student.

In comparison with previous works, our method first proposes generating synthetic data at a lower resolution, guided by their class activation maps, to enhance image quality (Section \ref{sec:cam}). Next, we introduce two techniques to further improve the diversity of our models (Section \ref{sec:diver}). Finally, the overall process is summarized in Section \ref{sec:overall}.

\subsection{Lower-resolution Data-free Generation}
\label{sec:cam}

A major limitation of previous approaches is their generation of synthetic images at high resolutions ($224 \times 224$) without incorporating information from real images. This leads to images with significant noise, lacking the class-specific features essential for effective knowledge transfer, as illustrated in Figure \ref{fig:diff} (a). To address these limitations, we propose generating synthetic images at lower resolutions. 
\begin{equation}
    \hat{\vx} = \mathcal{G}_{e \times e}(\mathcal{Z}(\vf_{\vy})),
\end{equation}
where $\hat{\vx} \in \mathbb{R}^{e \times e}$ and $e$ is the target resolution (i.e., $e \ll l$).
\noindent
\textbf{CAM-Enhanced Quality Loss.} The important shape or pattern in sub-regions \cite{vit,if1,shapeformer,ppsn,pisd} is very important for classification, To ensure that the synthetic images $\hat{\vx}$ capture important information, we propose maximizing their CAM with the target map, which contains high values of class activation. First, we use the classic CAM method \cite{cam} to generate the matrix $M(\hat{\vx}, \hat{\vy})$ for the image $\hat{\vx}$ and class $\hat{\vy}$:
\begin{equation}
    M(\hat{\vx}, \hat{\vy}) = \sum_k \vw_k^{\hat{\vy}} \mathcal{T}_k(\hat{\vx}, \hat{\vy}),
\end{equation}
where $\vw_k^{\hat{\vy}}$ is the $k^{\text{th}}$ weight in the final classification head for class $\hat{\vy}$, and $\mathcal{T}_k$ is the $k^{\text{th}}$ feature map in the final layers of the model. Note that we only use the latent matrix of CAM, which is before the normalization and interpolation into full-resolution images. Then, the loss function $\mathcal{L}_{\mathcal{G}}$ is modified as follows:
\begin{align}
    \mathcal{L}_{\mathcal{G}} &= \alpha_{ce}\mathcal{L}_{ce} + \alpha_{adv}\mathcal{L}_{adv} + \alpha_{bn}\mathcal{L}_{bn} + \alpha_{cam}\mathcal{L}_{cam} \nonumber \\ 
    \mathcal{L}_{cam} &= \max\{0, M_{\text{target}} - M(\hat{\vx}, \hat{\vy})\}.
\end{align}
In this context, $M_{\text{target}}$ is a predefined mask with high values at the center and lower values at the borders, guiding the generator to produce the desired activation map $M(\hat{\boldsymbol{x}}, \hat{\boldsymbol{y}})$. By using a margin loss to define $\mathcal{L}_{cam}$, we encourage the values in $M(\hat{\boldsymbol{x}}, \hat{\boldsymbol{y}})$ to \textit{only sufficiently exceed} those in $M_{\text{target}}$, avoiding excessively high values that could negatively impact image quality while concentrating the important values of $M(\hat{\boldsymbol{x}}, \hat{\boldsymbol{y}})$ near the center. Furthermore, we conducted an ablation study in \textbf{Supplementary Material} to identify the most suitable mask for this task.

Thanks to the use of CAM-enhanced quality loss and lower-resolution images, as shown in Figure \ref{fig:diff} (c), generating synthetic images at lower resolutions improves accuracy by enabling the generator to capture critical features more effectively. Figure \ref{fig:moti1} further illustrates the substantial reduction in training time, highlighting the efficiency gains of this approach. Together, these findings underscore the advantages of low-resolution synthetic images in enhancing both performance and computational efficiency in DFKD for large-scale datasets.

\subsection{Improved Model Diversity}
\label{sec:diver}
\begin{figure*}[t]
\begin{center}
\includegraphics[width=\linewidth]{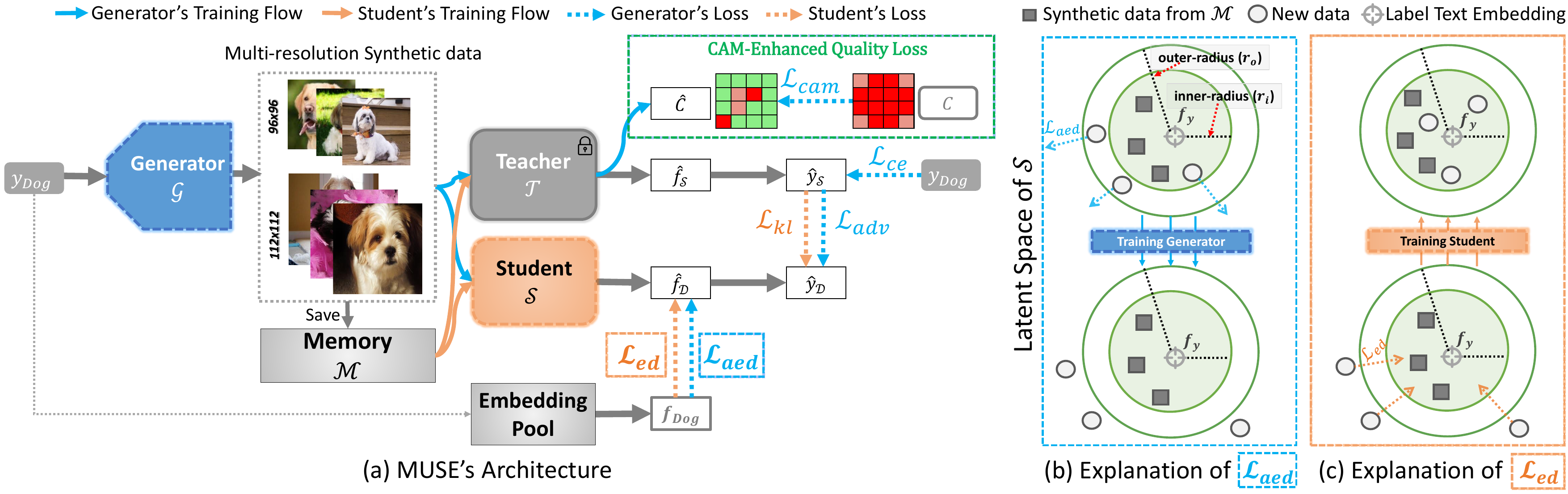}
\end{center}
\caption{(a) Overview of the MUSE architecture, illustrating the two-phase training process: generator training and student training. The model generates lower-resolution images and enhances their quality using CAM-Enhanced Quality Loss, while also promoting diversity through Embedding Diversity Loss ($\mathcal{L}_{ed}$ and $\mathcal{L}_{aed}$). (b) $\mathcal{L}_{ed}$ (Eq. \ref{eq:le}) aims to learn the embedding in $\mathcal{S}$ of all old data, bringing it closer to $\vf_{\vy}$, while (c) $\mathcal{L}_{aed}$ (Eq. \ref{eq:laed}) guides the generator $\mathcal{G}$ to produce new data that is distant from $\vf_{\vy}$, thus enhancing the model's diversity.}
\label{fig:arch}
\end{figure*}

While lower-resolution synthetic images enhance computational efficiency, they can also limit the model's ability to capture diverse and detailed features, as lower resolutions reduce the space available for representing such diversity.

\noindent
\textbf{Multi-resolution Data Generation.} To overcome this challenge, we propose a multi-resolution generation strategy that synthesizes images at various resolutions, effectively capturing both coarse and fine-grained features. Given a set of resolutions $E$, the synthetic data $\hat{\vx}$ is generated from each resolution $e \sim E$:
\begin{equation}
    \hat{\vx} = \mathcal{G}_{e \times e, e \in E}(\mathcal{Z}(\vf_{\vy})),
\end{equation}
\noindent
\textbf{Embedding Diversity.} Additionally, we introduce embedding diversity techniques to preserve distinct representations within the latent space, ensuring that rich feature representations are maintained even at lower resolutions. These techniques consist of two loss functions, which are used for training the generator $\mathcal{G}$ and the student $\mathcal{S}$, respectively.
 
In the student training phase, given a pool of synthetic data $\hat{\vx} \sim \mathcal{M}$, the student network $\mathcal{S}$ is trained using the following loss function:
\begin{align}
\label{eq:ls}
    \mathcal{L}_{\mathcal{S}} &= \mathcal{L}_{kl} + \alpha_{ed}\mathcal{L}_{ed}, \\
    \mathcal{L}_{ed} &= B(MSE(\hat{\vf}_{\mathcal{S}}, \vf_{\vy}), r_i), \\
    &= \max\{0, MSE(\hat{\vf}_{\mathcal{S}}, \vf_{\vy}) - r_i\},
    \label{eq:le}
\end{align}
where $\alpha_{ed}$ is a scaling factor, $\mathcal{L}_{kl}$ is computed by Eq. \ref{eq:kl}, $\hat{\vf}_{\mathcal{S}}$ is the latent embedding of $\hat{\vx}$ in the student model $\mathcal{S}$, and $\vf_{\vy}$ is the class-specific embedding representative. The purpose of the bounding term is to learn embeddings from the synthetic data pool $\mathcal{M}$ that are close to the class representative embedding $\vf_{\vy}$ of the original data. Inspired by \cite{lander}, we use the Bounding Loss (B Loss) to encourage the embedding of $\hat{\vy}$ to stay within an inner radius $r_i$, while preserving its intrinsic distance characteristics.

In the generator training phase, on the other hand, the generator aims to produce a new batch of synthetic data that is positioned far from the class embedding $\vf_{\vy}$. Similar to $\mathcal{L}_{cam}$, we apply a margin loss to ensure that the embedding of $\hat{\vx}$ in the teacher model $\mathcal{T}$ does not deviate excessively from the desired distribution.
\begin{align}
\label{eq:lg}
    \mathcal{L}_{\mathcal{G}} &= \alpha_{ce}\mathcal{L}_{ce} + \alpha_{adv}\mathcal{L}_{adv} + \alpha_{bn}\mathcal{L}_{bn} \nonumber\\
    &+  \alpha_{cam}\mathcal{L}_{cam} + \alpha_{aed}\mathcal{L}_{aed} \\
    \mathcal{L}_{aed} &= max\{0, r_o - MSE(\hat{\vf}_{\mathcal{S}}, \vf_{\vy})\}
    \label{eq:laed}
\end{align}
where $r_o > r_i$ represents the outer radius, and $\alpha$ are scaling parameters. 

We now explain how the cooperation between the generator and student in the \textit{embedding in-out game}, achieved by minimizing $\mathcal{L}_\mathcal{S}$ and $\mathcal{L}_\mathcal{G}$, promotes embedding diversity. Specifically, by minimizing $\mathcal{L}_{ed}$ during student training, the model learns to keep the latent embeddings of all previous data within an inner radius around $\vf_{\vy}$, positioning them closer to $\vf_{\vy}$ (Figure \ref{fig:arch} (b)). In contrast, $\mathcal{L}_{aed}$ guides the generator $\mathcal{G}$ to produce new data with latent embeddings that are distant from $\vf_{\vy}$ (Figure \ref{fig:arch} (c)). This setup encourages the new data to differ from the old data in latent space, thereby enhancing the diversity of the latent embeddings. 

\noindent
\textbf{Choosing Class Representative Embedding $\vf_{\vy}$.} The embedding $\vf_{\vy}$ plays a crucial role in promoting embedding diversity, and we consider two options for selecting $\vf_{\vy}$. First, since we use the generator from NAYER \cite{nayer} as our baseline, we propose using the label text embedding as $\vf_{\vy}$. Second, when the label text embedding is unavailable, we use the mean of the embeddings in $\mathcal{T}$ from the first batch as $\vf_{\vy}$. Both options serve as representative embeddings for the class. We conducted an ablation study (\textbf{Supplementary Material}) showing that both methods are comparable, with the label text embedding yielding slightly better performance.

\subsection{Overall Architecture}
\label{sec:overall}

The overall architecture of MUSE is shown in Figure \ref{fig:arch}, and the pseudo code can be found in Algorithm \ref{alg:muse}. First, MUSE embeds all label text using a text encoder. Then, our method undergoes training for $\mathcal{E}$ epochs. Each epoch consists of two distinct phases:
\begin{enumerate}[noitemsep,nolistsep]
    \item[(i)] The first phase involves training the generator. In each iteration $I$, as described in Algorithm \ref{alg:muse}, the noisy layer $\mathcal{Z}$ is reinitialized (line 5) before being used to learn the label text embedding $\vf_{\vy}$. The generator and noisy layer are then trained over $g$ steps using Eq. \ref{eq:lg} to optimize their performance (line 9).
    \item[(ii)] The second phase involves training the student network. During this phase, all generated samples are stored in the memory module $\mathcal{M}$ to mitigate the risk of forgetting (line 10), following a similar approach as outlined in \cite{fastdfkd}. Finally, the student model is trained using Eq. \ref{eq:ls} over $S$ iterations, utilizing samples from $\mathcal{M}$ (lines 12 and 13).
\end{enumerate}

\begin{algorithm}[t]
\caption{MUSE}
\label{alg:muse}
\kwInput{pre-trained teacher $\mathcal{T}_{\theta_\mathcal{T}}$, student $\mathcal{S}_{\theta_\mathcal{S}}$, generator $\mathcal{G}_{\theta_\mathcal{G}}$, text encoder $\mathcal{C}_{\theta_\mathcal{C}}$, list of labels ${\vy}$ and list of text of these labels $Y_{{\vy}}$\; }
\kwOutput{An optimized student  $\mathcal{S}_{\theta_\mathcal{S}}$}
Initializing $\mathcal{P} = \{\}, \mathcal{M} = \{\}$\;
Store all embeddings $\vf_{{\vy}} = \mathcal{C}(Y_{{\vy}})$ into $\mathcal{P}$\;
\For{$\mathcal{E}$ \textit{epochs}}{
    \For{$I$ \textit{iterations}}{
        Randomly reinitializing noisy layers $\mathcal{Z}_{\theta_{\mathcal{Z}}}$ and pseudo label $\hat{\vy}$ for each iteration\;
        Query $\vf_{\hat{\vy}} \sim \mathcal{P}$\;
        \For{$g$ \textit{steps}}{
            ${\hat{\vx}} = {\mathcal{G}}_{e\times e, e \in E}(\mathcal{Z}(\vf_{\vy}))$\;
            Update $\theta_{\mathcal{G}}, \theta_{\mathcal{Z}}$ by minimizing $\mathcal{L}_{\mathcal{G}}$ Eq. \ref{eq:lg};
        }
        $\mathcal{M} \gets \mathcal{M} \cup \hat{\vx}$\;
    }
    \For{$S$ \textit{iterations}}{
    $\hat{\vx} \sim \mathcal{M}$\;
    Update $\theta_{\mathcal{S}}$ by minimizing $\mathcal{L}_\mathcal{S}$ (Eq. \ref{eq:ls});
    }
}
\end{algorithm}

\begin{table*}[hpt]
\centering
\begin{adjustbox}{width=\linewidth}

\begin{tabular}{@{}l|cccc|cccc|cccc@{}}
\toprule
Dataset  & \multicolumn{4}{c}{Imagenetee}                                                                & \multicolumn{4}{c}{Imagewoof}                                                                 & \multicolumn{4}{c}{ImageNet1k}                                                                   \\ \midrule
Teacher  & \multicolumn{4}{c}{ResNet50 (92.86)}                                                          & \multicolumn{4}{c}{ResNet50 (86.84)}                                                          & \multicolumn{4}{c}{ResNet50 (80.86)}                                                             \\
Student  & \multicolumn{4}{c}{MobileNetV2 (90.42)}                                                       & \multicolumn{4}{c}{MobileNetV2 (82.69)}                                                       & \multicolumn{4}{c}{MobileNetV2 (71.88)}                                                          \\
\midrule
Data Ratio    & 1\%                  & 5\%                  & 10\%                   & 20\%    & 1\%                  & 5\%                  & 10\%                   & 20\%                       & 1\%                  & 5\%                  & 10\%                   & 20\%                    \\
Fast100 \cite{fastdfkd}        & 8.92 (0.5h)           & 29.18 (0.5h)          & 39.12 (0.8h)          & 51.43 (1.4h)          & 5.42 (0.3h)           & 15.11 (0.5h)          & 23.45 (0.8h)          & 38.92 (1.4h)          & 4.78 (6.3h)           & 16.58 (23.5h)          & 22.12 (39.2h)          & 25.25 (65.6h)          \\
NAYER \cite{nayer}   & 9.54 (0.5h)           & 31.28 (0.5h)          & 42.24 (0.8h)          & 54.26 (1.4h)          & 6.99 (0.3h)           & 16.72 (0.5h)          & 27.43 (0.8h)          & 40.21 (1.4h)          & 6.32 (6.3h)           & 19.78 (23.5h)          & 25.43 (39.2h)          & 28.12 (65.6h)          \\
MUSE-S     & {\ul 35.32 (0.5h)}    & {\ul 80.11 (0.5h)}    & {\ul 87.21 (0.8h)}    & {\ul 88.53 (1.4h)}    & {\ul 21.25 (0.3h)}    & {\ul 36.24 (0.5h)}    & {\ul 71.42 (0.8h)}    & {\ul 74.53 (1.4h)}    & {\ul 22.41 (6.3h)}    & {\ul 40.63 (23.5h)}    & {\ul 46.25 (39.2h)}    & {\ul 53.24 (65.6h)}    \\

MUSE & \textbf{36.16 (0.8h)} & \textbf{81.21 (0.8h)} & \textbf{88.12 (1.2h)} & \textbf{89.21 (2.1h)} & \textbf{22.43 (0.5h)} & \textbf{37.51 (0.8h)} & \textbf{72.11 (1.2h)} & \textbf{75.12 (2.1h)} & \textbf{24.25 (9.3h)} & \textbf{42.24 (30.1h)} & \textbf{47.12 (58.5h)} & \textbf{54.41 (80.4h)} \\
\midrule
Dataset  & \multicolumn{4}{c}{Imagenetee}                                                                & \multicolumn{4}{c}{Imagewoof}                                                                 & \multicolumn{4}{c}{ImageNet1k}                                                                   \\
\midrule
Teacher  & \multicolumn{4}{c}{Resnet34 (94.06)}                                                          & \multicolumn{4}{c}{Resnet34 (83.02)}                                                          & \multicolumn{4}{c}{Resnet34 (73.31)}                                                             \\
Student  & \multicolumn{4}{c}{Resnet18 (93.53)}                                                          & \multicolumn{4}{c}{Resnet18 (82.59)}                                                          & \multicolumn{4}{c}{Resnet18 (69.76)}                                                             \\
\midrule
Data Ratio    & 1\%                  & 5\%                  & 10\%                   & 20\%    & 1\%                  & 5\%                  & 10\%                   & 20\%                       & 1\%                  & 5\%                  & 10\%                   & 20\%                    \\
Fast100 \cite{fastdfkd}    & 8.51 (0.3h)           & 28.32 (0.5h)          & 38.25 (0.6h)          & 49.11 (1.0h)            & 5.21 (0.3h)           & 14.24 (0.5h)          & 23.54 (0.8h)          & 35.72 (1.1h)          & 3.63 (5.1h)           & 15.52 (18.8h)          & 20.12 (31.4h)          & 25.96 (52.5h)          \\
NAYER \cite{nayer}   & 9.35 (0.3h)           & 32.17 (0.5h)          & 42.57 (0.6h)          & 52.72 (1.0h)            & 6.72 (0.3h)           & 15.62 (0.5h)          & 25.27 (0.8h)          & 38.25 (1.1h)          & 5.81 (5.1h)           & 18.86 (18.8h)          & 23.98 (31.4h)          & 28.11 (52.5h)          \\
MUSE-S     & {\ul 34.52 (0.3h)}    & {\ul 80.32 (0.5h)}    & {\ul 86.67 (0.6h)}    & {\ul 88.25 (1.0h)}      & {\ul 20.52 (0.3h)}    & {\ul 36.25 (0.5h)}    & {\ul 59.85 (0.8h)}    & {\ul 73.74 (1.1h)}    & {\ul 22.32 (5.1h)}    & {\ul 40.82 (18.8h)}          & {\ul45.95 (31.4h)}          & {\ul 53.96 (52.5h)}    \\
MUSE & \textbf{35.21 (0.5h)} & \textbf{82.21 (0.8h)} & \textbf{87.21 (1.1h)} & \textbf{88.72 (1.5h)} & \textbf{21.12 (0.5h)} & \textbf{37.31 (0.8h)} & \textbf{60.04 (1.2h)} & \textbf{74.52 (1.5h)} & \textbf{24.16 (7.5h)} & \textbf{42.84 (24.1h)} & \textbf{47.13 (46.8h)} & \textbf{54.98 (64.3h)} \\ 
\bottomrule
\end{tabular}

\end{adjustbox}
\caption{Distillation results of our MUSE (multi-resolution)  and MUSE-S (single-resolution) compared to the current state-of-the-art DFKD methods, including NAYER \cite{nayer} and popular Fast methods \cite{fastdfkd} (Fast100 refers to using 100 generation steps for Fast), across multiple datasets (Imagenette, Imagewoof, and ImageNet1k) at different data ratios. The methods were evaluated on two popular knowledge distillation pairs: (ResNet50 to MobileNetV2) and (ResNet34 to ResNet18). Bold numbers indicate the highest accuracy, while underlined numbers indicate the second-highest. All experiments were conducted across 3 runs, and the mean accuracy is reported.}
\label{tab:lsdataset}
\end{table*}

\section{Experiments}
\subsection{Experimental Settings}

For large-scale datasets, we evaluated our method using two popular pairs of backbones: ResNet34/ResNet18 \cite{resnet} and ResNet50/MobileNetV2 \cite{mnv2} on three widely-used benchmarks: Tiny-ImageNet \cite{tin},  ImageNet1k \cite{imagenet}, which includes 1,000 object categories and over 1.2 million labeled training images, along with its subsets Imagenette and Imagewoof, each containing 10 sub-classes. For small-scale datasets, we conducted experiments with ResNet \cite{resnet}, VGG \cite{vgg}, and WideResNet (WRN) \cite{wrn} across CIFAR-10 and CIFAR-100 \cite{c10}. Both CIFAR-10 and CIFAR-100 contain 60,000 images (50,000 for training and 10,000 for testing), with 10 and 100 categories, respectively, and all images have a resolution of $32 \times 32$ pixels.

All comparative experiments were conducted using the AdamW optimizer with an initial learning rate of 0.0001 and the LambdaLR scheduler. We fixed the following hyperparameters for all experiments: $\alpha_{ce}$ at 0.5, $\alpha_{bn}$ at 10, $\alpha_{adv}$ at 1.3 (as in \cite{nayer}), $\alpha_{ed}$ at 10, $\alpha_{aed}$ at 5, $\alpha_{cam}$ at 0.1, $r_i$ at 0.015, and $r_o$ at 0.03. All runs were executed on a single NVIDIA A100 40GB GPU with four DataLoader workers. Additional details on the architectures, parameter settings, parameter sensitivity and further analysis can be found in the \textbf{Supplementary Material}.

\begin{table*}[t]
\centering
\begin{adjustbox}{width=\linewidth}

\begin{tabular}{@{}l|ccccc|ccccc@{}}
\toprule
\textbf{Dataset}                 & \multicolumn{5}{c}{\textbf{CIFAR10}}                                                                                      & \multicolumn{5}{c}{\textbf{CIFAR100}}                                                                                             \\ \midrule
\multirow{2}{*}{\textbf{Method}} & \textbf{R34}          & \textbf{V11}        & \textbf{W402}         & \textbf{W402}               & \textbf{W402}         & \textbf{R34}            & \textbf{V11}                & \textbf{W402}         & \textbf{W402}         & \textbf{W402}             \\
\multicolumn{1}{c}{}                                 & \textbf{R18}          & \textbf{R18}        & \textbf{W161}         & \textbf{W401}               & \textbf{W162}         & \textbf{R18}            & \textbf{R18}                & \textbf{W161}         & \textbf{W401}         & \textbf{W162}             \\  \midrule
Teacher                                              & 95.70                 & 92.25               & 94.87                 & 94.87                 & 94.87                 & 77.94                   & 71.32                 & 75.83                 & 75.83                 & 75.83               \\
Student                                              & 95.20                 & 95.20               & 91.12                 & 93.94                 & 93.95                 & 77.10                   & 77.10                 & 65.31                 & 72.19                 & 73.56               \\
KD                                                   & 95.20                 & 95.20               & 95.20                 & 95.20                 & 95.20                 & 77.87                   & 75.07                 & 64.06                 & 68.58                 & 70.79               \\ \midrule
DeepInv \cite{adi}                                            & 93.26 (25.6h)         & 90.36 (13.2h)       & 83.04 (10.6h)         & 86.85 (13.8h)         & 89.72 (11.7h)         & 61.32 (25.9h)           & 54.13 (13.1h)         & 53.77 (10.6h)         & 61.33 (13.9h)         & 61.34 (11.7h)       \\
DFQ \cite{dfq}                                       & 94.61 (152.2h)        & 90.84 (104.9h)      & 86.14 (48.4h)         & 91.69 (78.8h)         & 92.01 (118.5h)        & {77.01 (151.2h)} & 66.14 (105.1h)        & 51.27 (48.5h)         & 54.43 (78.8h)         & 64.79 (59.5h)       \\
ZSKT \cite{zskt}                                             & 93.32 (304.1h)        & 89.46 (209.6h)      & 83.74 (96.7h)         & 86.07 (157.5h)         & 89.66 (118.8h)         & 67.74 (304.1h)           & 54.31 (209.6h)         & 36.66 (96.8h)          & 53.60 (157.6h)         & 53.60(157.6h)       \\
Fast5 \cite{fastdfkd}                                     & 93.63 (4.3h)          & 89.94 (3.2h)        & 88.90 (1.6h)           & 92.04 (2.4h)          & 91.96 (1.9h)          & 72.82 (4.4h)            & 65.28 (3.1h)          & 52.90 (1.6h)           & 61.80 (2.4h)           & 63.83 (1.9h)        \\
Fast10 \cite{fastdfkd}                                     & 94.05 (4.7h)          & 90.53 (3.2h)        & 89.29 (1.7h)          & 92.51 (2.5h)          & 92.45 (2.1h)          & 74.34 (4.5h)            & 67.44 (3.1h)          & 54.02 (1.7h)          & 63.91 (2.4h)          & 65.12 (2.0h)          \\
SS \cite{ss}                                        & 94.26 (4.2h)          & 90.67 (3.0h)          & 89.96 (1.5h)          & 93.23 (2.3h)          & 93.11 (1.8h)          & 75.16 (4.2h)            & \textbf{68.77 (2.9h)} & 55.61 (1.5h)          & 64.57 (2.3h)          & 65.28 (1.8h)        \\
NAYER \cite{nayer}                                      & 94.14 (4.8h)          & 89.81 (3.3h)        & 89.99 (1.8h)          & 92.91 (2.6h)          & 93.12 (2.1h)          & 74.15 (4.6h)            & 63.47 (3.2h)          & 57.12 (1.7h)          & 65.67 (2.5h)          & 67.41 (2.0h)          \\
MUSE                                                 & {\ul 94.31 (4.8h)}    & {\ul 90.51 (3.3h)}  & {\ul 90.67 (1.8h)}    & {\ul 93.41 (2.5h)}    & {\ul 93.25 (2.1h)}    & \underline{75.18 (4.6h)}            & {65.72 (3.3h)}          & {\ul 57.23 (1.7h)}    & {\ul 66.45 (2.5h)}    & {\ul 67.68 (2.0h)}    \\
MUSE-Mul                                             & \textbf{94.66 (7.3h)} & \textbf{90.92 (5.0h)} & \textbf{91.02 (2.7h)} & \textbf{93.57 (3.8h)} & \textbf{93.37 (3.2h)} & \textbf{75.21 (6.9h)}      & {\ul 66.21 (4.9h)}    & \textbf{58.81 (2.6h)} & \textbf{66.91 (3.8h)} & \textbf{68.47 (3.0h)}
\\ \bottomrule
\end{tabular}
\end{adjustbox}
\caption{The distillation results for the CIFAR10 and CIFAR100 datasets compare various methods, following the setup of \cite{ss}, with the data ratio set at 100\% (equivalent to 50,000 samples). The table presents the accuracy achieved by different student models with various architectures, such as ResNet (R) \cite{resnet}, VGG (V) \cite{vgg}, and WideResNet (W) \cite{wrn}, along with the corresponding training times in hours. We have re-executed the results from NAYER \cite{nayer} and compiled all other comparison results from \cite{ss}. To ensure a fair comparison with the findings in \cite{ss}, we also used the same NVIDIA V100 GPU for calculating the training time.}
\label{tab:ssdataset}
\end{table*}

\subsection{Data Scale Ratio}

Training models on the large-scale ImageNet dataset presents significant computational challenges \cite{adi,nayer}, especially when generating synthetic data. To address this, we set a limit on the amount of synthetic data produced for training the student model \cite{ss}. This approach aligns with practices in domains such as Continual Learning \cite{dfkd_cl1,dfkd_cl2} and Federated Learning \cite{lander,dfkd_fl1}, where synthetic data generation is also restricted to manage resource demands. In our study, we test various data ratios on ImageNet1k and its subsets to provide a balanced comparison across methods. For instance, using a 10\% data ratio on ImageNet results in roughly 100,000 samples at a $224\times 224$ resolution.


\noindent
\textbf{Data Scale Ratio for Lower-Resolution Images.} In this work, we propose generating lower-resolution data for training the student model, which reduces both memory requirements for data storage and computational time for model training. Specifically, storing and training with one $224 \times 224$ image requires roughly the same resources as using four $112 \times 112$ images or five $96 \times 96$ images. This enables our MUSE framework to produce a larger number of training samples while maintaining the same data scale ratio. For example, at a data ratio of 10\%, previous methods generate 10,000 samples for ImageNet1k, whereas our MUSE-S (using $112 \times 112$ images) can generate 40,000 samples without increasing training time or memory storage.

For multi-resolution data generation, in theory, MUSE with multi-resolution data (e.g., 25,000 $96 \times 96$ images and 20,000 $112 \times 112$ images) maintains the same computational complexity as using 40,000 $112 \times 112$ images. However, due to current limitations in PyTorch, multi-resolution MUSE runs more slowly. To ensure fair comparisons across all experiments, we present two versions of our method: MUSE (multi-resolution generation) and MUSE-S (single-resolution variant). Details on the resolutions used in MUSE and MUSE-S for each dataset are provided in the \textbf{Supplementary Material}.


\subsection{Results and Analysis}

\noindent
\textbf{Comparison on Large-scale and High-resolution Datasets.} Table \ref{tab:lsdataset} presents the distillation results across multiple datasets, including Imagenette, Imagewoof, and ImageNet1k, comparing the performance of MUSE-S and MUSE with existing methods such as Fast100 \cite{fastdfkd} and NAYER \cite{nayer} at varying data ratios. For example, with the ResNet50 to MobileNetV2 pair at a 10\% data ratio on the Imagenette dataset, MUSE-S outperforms both Fast and NAYER, achieving an accuracy of 87.21\%, compared to 39.12\% and 42.24\%, respectively. The multi-resolution version, MUSE, further improves this performance to 88.12\%. Similarly, on the ImageNet1k dataset, MUSE-S achieves an accuracy of 45.95\% at a 10\% data ratio with the ResNet34 to ResNet18 pair, surpassing Fast (20.12\%) and NAYER (23.98\%), while MUSE increases the accuracy to 47.13\%. These results demonstrate that MUSE-S, particularly with multi-resolution generation (MUSE), significantly enhances knowledge transfer efficiency and accuracy, while maintaining similar (MUSE-S) or slightly higher (MUSE) computational time.

\noindent
\textbf{Comparison on Small-scale Datasets.} Table \ref{tab:ssdataset} presents a comparison of the distillation performance of various methods on the CIFAR10 and CIFAR100 datasets, showcasing the effectiveness of both MUSE-S and its multi-resolution variant, MUSE. For instance, on CIFAR10, applying MUSE to ResNet34 and ResNet18 achieves an accuracy of 94.51\% with a training time of just 7.3 hours, surpassing methods like KD and DeepInv, which achieve accuracies of 95.20\% and 93.26\%, respectively, but at significantly higher computational costs. Similarly, on CIFAR100, MUSE achieves the highest accuracy of 75.21\%, outperforming other approaches such as NAYER (74.15\%) and DFQ (77.01\%), while maintaining a reasonable training time. These results highlight MUSE's ability to deliver superior performance while efficiently utilizing computational resources across both datasets.

\begin{figure}[t]
\begin{center}
\includegraphics[width=\linewidth]{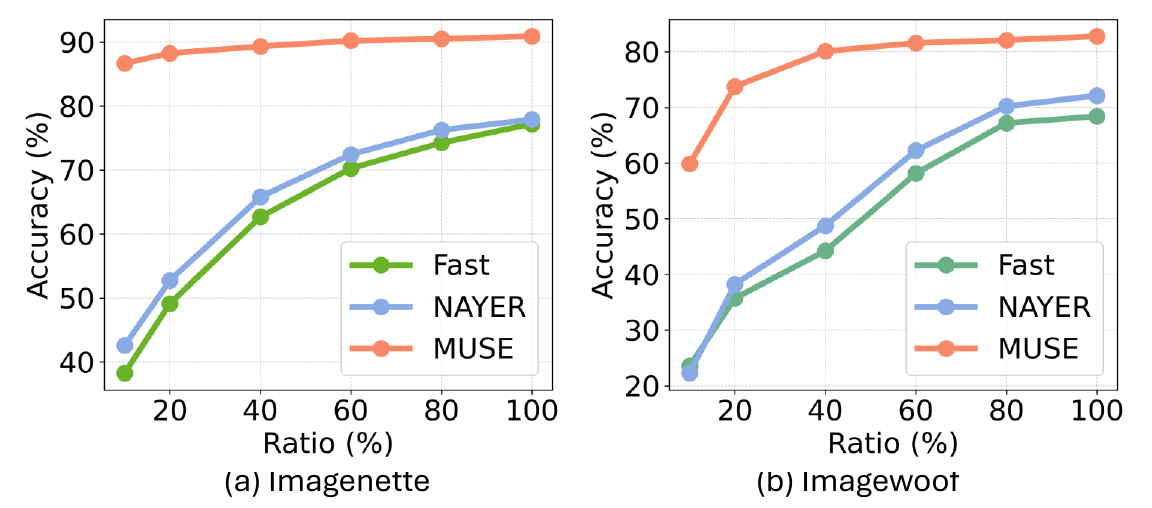}
\end{center}
\caption{The accuracy at data ratios from 10\% to 100\% is shown for the teacher (ResNet34) and student (ResNet18) models.}
\label{fig:higher_ratio}
\end{figure}

\noindent
\textbf{Comparasion in Higher Data Ratios Setting.} To further demonstrate the benefits of our methods, we also conducted experiments on higher data ratio settings, as shown in Figure \ref{fig:higher_ratio}. The results indicate that our methods achieve higher accuracy across all ratio settings on both the Imagenette and Imagewoof datasets. Particularly at lower ratios, the difference is significant. For example, at a ratio of 20\% on Imagenette, our MUSE method achieves an accuracy approximately 40\% higher than the compared methods. These results demonstrate the effectiveness of our models.

\subsection{Ablation Study}

\noindent
\textbf{Effectiveness of Lower-resolution.} In Table \ref{tab:lcam}, we present the accuracy of our methods with different image resolutions. The results clearly show that the accuracy of models decreases significantly when the resolution is either too small ($64 \times 64$) or too large ($224 \times 224$), with the highest accuracy achieved at $96 \times 96$. This illustrates the importance of selecting an appropriate resolution for synthetic data, balancing both computational efficiency and model performance.

\begin{table}[t]
\centering
\begin{adjustbox}{width=\linewidth}
\begin{tabular}{@{}lcccccccc@{}}
\toprule
Resolution ($R\times R$)           & 224   & 192   & 144   & 128   & 112   & 96    & 80    & 64    \\ \midrule
With $\mathcal{L}_{cam}$    & 37.27 & 40.65 & 65.25 & 70.21 & 78.21 & 80.32 & 77.21 & 40.21 \\
Without $\mathcal{L}_{cam}$ & 32.17 & 34.26 & 58.21 & 65.21 & 72.25 & 75.12 & 71.23 & 34.91 \\ \bottomrule
\end{tabular}
\end{adjustbox}
\caption{Performance Comparison across multiple data resolutions in Imagenette (ResNet34/ResNet18 case) with the same Data Ratio at 5\%.}
\label{tab:lcam}
\end{table}

\noindent
\textbf{Effectiveness of CAM-Enhanced Quality Loss.} As shown in Table \ref{tab:lcam}, adding the CAM-enhanced quality loss term, $\mathcal{L}_{cam}$, improves performance, particularly at intermediate resolutions like $128 \times 128$ and $112 \times 112$. At these resolutions, the model achieves 70.21\% and 78.21\% accuracy, outperforming settings without $\mathcal{L}_{cam}$ by 5-6 percentage points, highlighting its effectiveness, especially at lower resolutions.

\noindent
\textbf{Effectiveness of Multi-resolution Data Generation.} Tables \ref{tab:lsdataset} and \ref{tab:ssdataset} demonstrate that MUSE, using multi-resolution data generation, outperforms other distillation methods in both accuracy and efficiency across various datasets. For instance, on CIFAR10, MUSE achieves 94.51\% accuracy, surpassing NAYER and SSD-KD. Similarly, on CIFAR100, MUSE reaches 75.21\%, outperforming MUSE-S and NAYER, while also delivering superior performance on Imagenette, showcasing its robustness.

\begin{figure}[t]
\begin{center}
\includegraphics[width=0.9\linewidth]{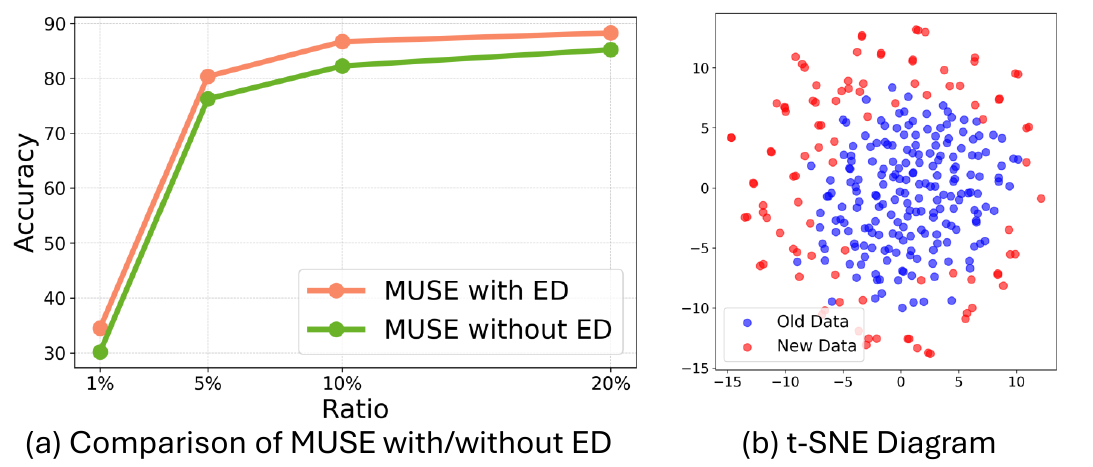}
\end{center}
\caption{(a) Accuracy of our MUSE method with and without Embedding Diversity (ED) for ResNet34 and ResNet18. (b) t-SNE visualization of the embeddings: synthetic data from the $\mathcal{M}$ pool (blue) and newly generated data (red).}
\label{fig:ed}
\end{figure}

\noindent
\textbf{Effectiveness of Embedding Diversity.} Figure \ref{fig:ed} shows that MUSE with Embedding Diversity consistently outperforms without ED at all data ratios, especially at lower ratios (1\% and 5\%), emphasizing ED's crucial role. Additionally, Figure \ref{fig:ed} (b) illustrates that new data typically occupies a distinct region in latent space, enhancing model diversity.

\subsection{Visualization}

\begin{figure}[t]
\begin{center}
\includegraphics[width=\linewidth]{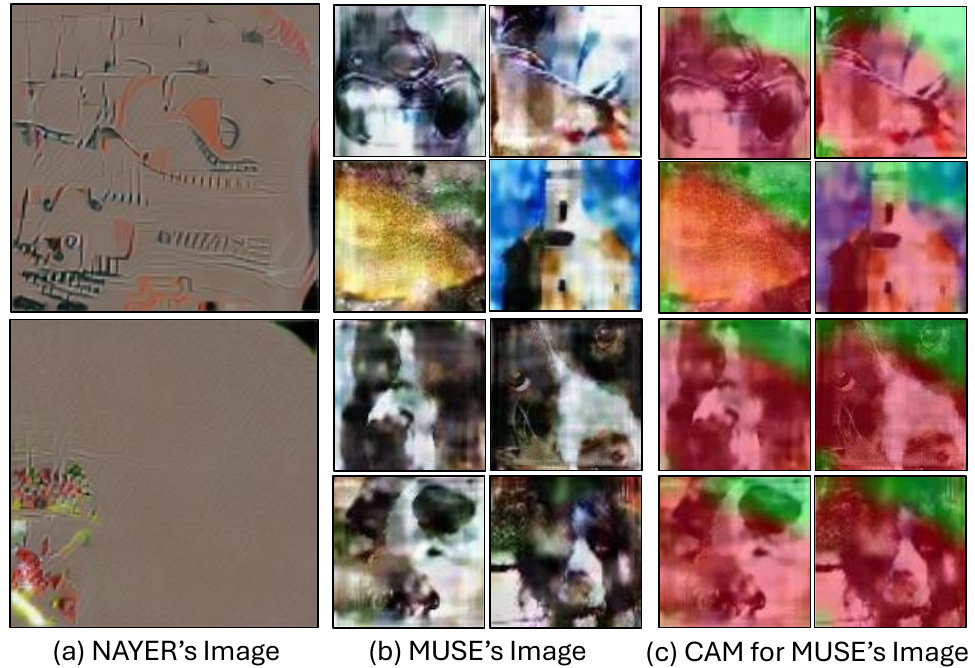}
\end{center}
\caption{(a-b) Synthetic data generated for ImageNet1k, with NAYER (at $224 \times 224$ resolution) and our MUSE (at $112 \times 112$ resolution). (c) Class activation map for our MUSE's images. Please note that the values of the class activation map are shown before normalization.}
\label{fig:visual}
\end{figure}

Figure \ref{fig:visual} shows synthetic images generated by NAYER (a) at $224 \times 224$ and MUSE (b) at $112 \times 112$, both after 100 generator training steps on ImageNet using ResNet-50 as the teacher. While challenging for human recognition and differing from real datasets, MUSE’s lower-resolution images capture key class-level features, showing superior quality over NAYER \cite{nayer}. In (c), the CAM for MUSE’s images reveals high CAM ratios across most pixels, highlighting the benefit of CAM-Enhanced Quality Loss.

\section{Limitation and Future works}

\noindent
\textbf{Lower-resolution for Vision Transformer.} A key challenge in our approach is training the student model with lower-resolution images, which are then tested on full-resolution data. This is particularly challenging for patch-based models, such as Vision Transformer (ViT) and its variants \cite{vit,deit}, that do not rely on CNN architectures. To address this, we propose reducing the number of patches input into the Vision Transformer. With the standard patch size of $16 \times 16$ used by ViT and our chosen image resolution of $112 \times 112$, we generate a $7 \times 7$ grid of patches instead of the original $14 \times 14$. By focusing on the center position embedding, our method, as shown in Table \ref{tab:vit}, outperforms the original NAYER training, achieving improvements of over two percentage points. Details of this technique are provided in the \textbf{Supplementary Material}.

\begin{table}[h!]
\centering
\begin{adjustbox}{width=0.9\linewidth}
\begin{tabular}{@{}lcccc@{}}
\toprule
Data Ratio  & \multicolumn{2}{c}{1\%}           & \multicolumn{2}{c}{5\%}           \\ \midrule
Metric & Top 1 Accuracy & Top 5 Accuracy & Top 1 Accuracy & Top 5 Accuracy \\
NAYER  & 4.52           & 19.45          & 16.24          & 43.24          \\
MUSE   & \textbf{15.24}          & \textbf{36.52}          & \textbf{28.24}          & \textbf{60.24}          \\ \bottomrule
\end{tabular}
\end{adjustbox}
\caption{Performance comparison between our MUSE and NAYER for knowledge transfer from DeiT-B (Teacher) to DeiT-Tiny (Student).}
\label{tab:vit}
\end{table}

\noindent
\textbf{Future Work.} Our paper employs a customized version of the classic CAM, designed to facilitate backpropagation in obtaining the activation matrix. This approach opens the door to exploring other techniques, such as Grad-CAM \cite{gradcam} or attention-based scores \cite{attentioncam}, to further enhance the task. Additionally, optimizing multi-resolution techniques for faster processing times presents another promising direction for improvement.

\section{Conclusion}
In this paper, we introduce MUSE, a novel approach to overcoming the challenges of traditional DFKD in large-scale datasets. MUSE generates synthetic images at lower resolutions and uses CAMs to preserve class-specific features, addressing noisy image generation and computational inefficiency, especially for large datasets like ImageNet1k. The use of multi-resolution generation and embedding diversity further improves the diversity of learned representations, enhancing student model performance. Experimental results demonstrate that MUSE achieves SOTA performance, with two-digit improvements on ImageNet1k.

{\small
\bibliographystyle{ieee_fullname}
\bibliography{egbib}

\begin{thebibliography}{10}\itemsep=-1pt

\bibitem{dfkd_gr}
Kuluhan Binici, Shivam Aggarwal, Nam~Trung Pham, Karianto Leman, and Tulika Mitra.
\newblock Robust and resource-efficient data-free knowledge distillation by generative pseudo replay.
\newblock In {\em Proceedings of the AAAI Conference on Artificial Intelligence}, volume~36, pages 6089--6096, 2022.

\bibitem{dfq}
Yoojin Choi, Jihwan Choi, Mostafa El-Khamy, and Jungwon Lee.
\newblock Data-free network quantization with adversarial knowledge distillation.
\newblock In {\em Proceedings of the IEEE/CVF Conference on Computer Vision and Pattern Recognition Workshops}, pages 710--711, 2020.

\bibitem{imagenet}
Jia Deng, Wei Dong, Richard Socher, Li-Jia Li, Kai Li, and Li Fei-Fei.
\newblock Imagenet: A large-scale hierarchical image database.
\newblock In {\em Proceedings of the IEEE Conference on Computer Vision and Pattern Recognition (CVPR)}, 2009.

\bibitem{mad}
Kien Do, Thai~Hung Le, Dung Nguyen, Dang Nguyen, Haripriya Harikumar, Truyen Tran, Santu Rana, and Svetha Venkatesh.
\newblock Momentum adversarial distillation: Handling large distribution shifts in data-free knowledge distillation.
\newblock {\em Advances in Neural Information Processing Systems}, 35:10055--10067, 2022.

\bibitem{vit}
Alexey Dosovitskiy.
\newblock An image is worth 16x16 words: Transformers for image recognition at scale.
\newblock {\em arXiv preprint arXiv:2010.11929}, 2020.

\bibitem{fastdfkd}
Gongfan Fang, Kanya Mo, Xinchao Wang, Jie Song, Shitao Bei, Haofei Zhang, and Mingli Song.
\newblock Up to 100x faster data-free knowledge distillation.
\newblock In {\em Proceedings of the AAAI Conference on Artificial Intelligence}, volume~36, pages 6597--6604, 2022.

\bibitem{cmi}
Gongfan Fang, Jie Song, Xinchao Wang, Chengchao Shen, Xingen Wang, and Mingli Song.
\newblock Contrastive model inversion for data-free knowledge distillation.
\newblock {\em arXiv preprint arXiv:2105.08584}, 2021.

\bibitem{resnet}
Kaiming He, Xiangyu Zhang, Shaoqing Ren, and Jian Sun.
\newblock Deep residual learning for image recognition.
\newblock In {\em Proceedings of the IEEE conference on computer vision and pattern recognition}, pages 770--778, 2016.

\bibitem{kd}
Geoffrey Hinton, Oriol Vinyals, and Jeff Dean.
\newblock Distilling the knowledge in a neural network.
\newblock {\em arXiv preprint arXiv:1503.02531}, 2015.

\bibitem{c10}
Alex Krizhevsky, Geoffrey Hinton, et~al.
\newblock Learning multiple layers of features from tiny images.
\newblock 2009.

\bibitem{shapeformer}
Xuan-May Le, Ling Luo, Uwe Aickelin, and Minh-Tuan Tran.
\newblock Shapeformer: Shapelet transformer for multivariate time series classification.
\newblock In {\em Proceedings of the 30th ACM SIGKDD Conference on Knowledge Discovery and Data Mining}, pages 1484--1494, 2024.

\bibitem{ppsn}
Xuan-May Le, Minh-Tuan Tran, and Van-Nam Huynh.
\newblock Learning perceptual position-aware shapelets for time series classification.
\newblock In {\em Joint European Conference on Machine Learning and Knowledge Discovery in Databases}, pages 53--69. Springer, 2022.

\bibitem{tin}
Ya Le and Xuan Yang.
\newblock Tiny imagenet visual recognition challenge.
\newblock {\em CS 231N}, 7(7):3, 2015.

\bibitem{attentioncam}
Saebom Leem and Hyunseok Seo.
\newblock Attention guided cam: Visual explanations of vision transformer guided by self-attention.
\newblock In {\em Proceedings of the AAAI Conference on Artificial Intelligence}, volume~38, pages 2956--2964, 2024.

\bibitem{dfkd_cl1}
Xiaorong Li, Shipeng Wang, Jian Sun, and Zongben Xu.
\newblock Memory efficient data-free distillation for continual learning.
\newblock {\em Pattern Recognition}, 144:109875, 2023.

\bibitem{dfkd_cl2}
Xiaorong Li, Shipeng Wang, Jian Sun, and Zongben Xu.
\newblock Variational data-free knowledge distillation for continual learning.
\newblock {\em IEEE Transactions on Pattern Analysis and Machine Intelligence}, 45(10):12618--12634, 2023.

\bibitem{ss}
He Liu, Yikai Wang, Huaping Liu, Fuchun Sun, and Anbang Yao.
\newblock Small scale data-free knowledge distillation.
\newblock In {\em Proceedings of the IEEE/CVF Conference on Computer Vision and Pattern Recognition}, pages 6008--6016, 2024.

\bibitem{zskt}
Paul Micaelli and Amos~J Storkey.
\newblock Zero-shot knowledge transfer via adversarial belief matching.
\newblock {\em Advances in Neural Information Processing Systems}, 32, 2019.

\bibitem{zskd}
Gaurav~Kumar Nayak, Konda~Reddy Mopuri, Vaisakh Shaj, Venkatesh~Babu Radhakrishnan, and Anirban Chakraborty.
\newblock Zero-shot knowledge distillation in deep networks.
\newblock In {\em International Conference on Machine Learning}, pages 4743--4751. PMLR, 2019.

\bibitem{kakr}
Gaurav Patel, Konda~Reddy Mopuri, and Qiang Qiu.
\newblock Learning to retain while acquiring: Combating distribution-shift in adversarial data-free knowledge distillation.
\newblock In {\em Proceedings of the IEEE/CVF Conference on Computer Vision and Pattern Recognition}, pages 7786--7794, 2023.

\bibitem{kd3}
Zengyu Qiu, Xinzhu Ma, Kunlin Yang, Chunya Liu, Jun Hou, Shuai Yi, and Wanli Ouyang.
\newblock Better teacher better student: Dynamic prior knowledge for knowledge distillation.
\newblock {\em arXiv preprint arXiv:2206.06067}, 2022.

\bibitem{mnv2}
Mark Sandler, Andrew Howard, Menglong Zhu, Andrey Zhmoginov, and Liang-Chieh Chen.
\newblock Mobilenetv2: Inverted residuals and linear bottlenecks.
\newblock In {\em Proceedings of the IEEE Conference on Computer Vision and Pattern Recognition (CVPR)}, pages 4510--4520, 2018.

\bibitem{nlpkd}
Victor Sanh, Lysandre Debut, Julien Chaumond, and Thomas Wolf.
\newblock Distilbert, a distilled version of bert: smaller, faster, cheaper and lighter.
\newblock {\em arXiv preprint arXiv:1910.01108}, 2019.

\bibitem{gradcam}
Ramprasaath~R Selvaraju, Michael Cogswell, Abhishek Das, Ramakrishna Vedantam, Devi Parikh, and Dhruv Batra.
\newblock Grad-cam: Visual explanations from deep networks via gradient-based localization.
\newblock In {\em Proceedings of the IEEE international conference on computer vision}, pages 618--626, 2017.

\bibitem{vgg}
Karen Simonyan and Andrew Zisserman.
\newblock Very deep convolutional networks for large-scale image recognition.
\newblock {\em arXiv preprint arXiv:1409.1556}, 2014.

\bibitem{deit}
Hugo Touvron, Matthieu Cord, Matthijs Douze, Francisco Massa, Alexandre Sablayrolles, and Herv{\'e} J{\'e}gou.
\newblock Training data-efficient image transformers \& distillation through attention.
\newblock In {\em International conference on machine learning}, pages 10347--10357. PMLR, 2021.

\bibitem{lander}
Minh-Tuan Tran, Trung Le, Xuan-May Le, Mehrtash Harandi, and Dinh Phung.
\newblock Text-enhanced data-free approach for federated class-incremental learning.
\newblock In {\em Proceedings of the IEEE/CVF Conference on Computer Vision and Pattern Recognition}, pages 23870--23880, 2024.

\bibitem{nayer}
Minh-Tuan Tran, Trung Le, Xuan-May Le, Mehrtash Harandi, Quan~Hung Tran, and Dinh Phung.
\newblock Nayer: Noisy layer data generation for efficient and effective data-free knowledge distillation.
\newblock In {\em Proceedings of the IEEE/CVF Conference on Computer Vision and Pattern Recognition}, pages 23860--23869, 2024.

\bibitem{pisd}
Minh-Tuan Tran, Xuan-May Le, Van-Nam Huynh, and Sung-Eui Yoon.
\newblock Pisd: A linear complexity distance beats dynamic time warping on time series classification and clustering.
\newblock {\em Engineering Applications of Artificial Intelligence}, 138:109222, 2024.

\bibitem{kd2}
Zhendong Yang, Zhe Li, Xiaohu Jiang, Yuan Gong, Zehuan Yuan, Danpei Zhao, and Chun Yuan.
\newblock Focal and global knowledge distillation for detectors.
\newblock In {\em Proceedings of the IEEE/CVF Conference on Computer Vision and Pattern Recognition}, pages 4643--4652, 2022.

\bibitem{adi}
Hongxu Yin, Pavlo Molchanov, Jose~M Alvarez, Zhizhong Li, Arun Mallya, Derek Hoiem, Niraj~K Jha, and Jan Kautz.
\newblock Dreaming to distill: Data-free knowledge transfer via deepinversion.
\newblock In {\em Proceedings of the IEEE/CVF Conference on Computer Vision and Pattern Recognition}, pages 8715--8724, 2020.

\bibitem{sre2l}
Zeyuan Yin, Eric Xing, and Zhiqiang Shen.
\newblock Squeeze, recover and relabel: Dataset condensation at imagenet scale from a new perspective.
\newblock {\em Advances in Neural Information Processing Systems}, 36, 2024.

\bibitem{skd}
Ji~Won Yoon, Hyeonseung Lee, Hyung~Yong Kim, Won~Ik Cho, and Nam~Soo Kim.
\newblock Tutornet: Towards flexible knowledge distillation for end-to-end speech recognition.
\newblock {\em IEEE/ACM Transactions on Audio, Speech, and Language Processing}, 29:1626--1638, 2021.

\bibitem{spshnet}
Shikang Yu, Jiachen Chen, Hu Han, and Shuqiang Jiang.
\newblock Data-free knowledge distillation via feature exchange and activation region constraint.
\newblock In {\em Proceedings of the IEEE/CVF Conference on Computer Vision and Pattern Recognition}, pages 24266--24275, 2023.

\bibitem{wrn}
Sergey Zagoruyko and Nikos Komodakis.
\newblock Wide residual networks.
\newblock {\em arXiv preprint arXiv:1605.07146}, 2016.

\bibitem{visual1}
MD Zeiler.
\newblock Visualizing and understanding convolutional networks.
\newblock In {\em European conference on computer vision/arXiv}, volume 1311, 2014.

\bibitem{if1}
Qi Zhang, Yifei Wang, and Yisen Wang.
\newblock Identifiable contrastive learning with automatic feature importance discovery.
\newblock {\em Advances in Neural Information Processing Systems}, 36, 2024.

\bibitem{kd1}
Borui Zhao, Quan Cui, Renjie Song, Yiyu Qiu, and Jiajun Liang.
\newblock Decoupled knowledge distillation.
\newblock In {\em Proceedings of the IEEE/CVF Conference on computer vision and pattern recognition}, pages 11953--11962, 2022.

\bibitem{cam}
Bolei Zhou, Aditya Khosla, Agata Lapedriza, Aude Oliva, and Antonio Torralba.
\newblock Learning deep features for discriminative localization.
\newblock In {\em Proceedings of the IEEE conference on computer vision and pattern recognition}, pages 2921--2929, 2016.

\bibitem{detr}
Xizhou Zhu, Weijie Su, Lewei Lu, Bin Li, Xiaogang Wang, and Jifeng Dai.
\newblock Deformable detr: Deformable transformers for end-to-end object detection.
\newblock {\em arXiv preprint arXiv:2010.04159}, 2020.

\bibitem{dfkd_fl1}
Zhuangdi Zhu, Junyuan Hong, and Jiayu Zhou.
\newblock Data-free knowledge distillation for heterogeneous federated learning.
\newblock In {\em International conference on machine learning}, pages 12878--12889. PMLR, 2021.

\end{thebibliography}
}
\pagebreak
\appendix
\section{Training Details}
In this section, we provide the details of model training for our methods, including Teacher Training, Generator, and Student Training.

\subsection{Teacher Model Training Details}

In this work, we utilized the pretrained ResNet-50 and ResNet-34 models from PyTorch, trained on ImageNet1k, and trained them from scratch on the ImageNette and ImageWoof datasets. For CIFAR-10/CIFAR-100, we employed pretrained ResNet-34 and WideResNet-40-2 teacher models from \cite{fastdfkd, nayer}. The teacher models were trained using the SGD optimizer with an initial learning rate of 0.1, momentum of 0.9, and weight decay of 5e-4, with a batch size of 128 for 200 epochs. The learning rate decay followed a cosine annealing schedule.

\subsection{Generator Training Details}

To ensure fair comparisons, we adopt the generator architecture outlined in \cite{fastdfkd, nayer} and the Noisy Layer (\code{BatchNorm1D, Linear}) as described in \cite{nayer} for all experiments. This architecture has been proven effective in prior work and provides a solid foundation for evaluating the performance of our model. The generator network is designed to learn rich feature representations while maintaining computational efficiency. The details of the generator architecture, including layer specifications and output sizes, are provided in Table \ref{tab:g_arch}. Additionally, we use the Adam optimizer with a learning rate of 4e-3 to optimize the generator, ensuring stable convergence during training.

\begin{table}[h]
\centering
\begin{adjustbox}{width=\linewidth}
\begin{tabular}{@{}ll@{}}
\toprule
Output          & \textbf{Size   Layers}                              \\ \midrule
1000            & \code{Input}                                        \\
$128 \times h/4 \times w/4$ & \code{Linear}                        \\
$128 \times h/4 \times w/4$ & \code{BatchNorm1D}                        \\
$128 \times h/4 \times w/4$ & \code{Reshape}                        \\
$128 \times h/2 \times w/2$ & \code{SpectralNorm (Conv (3 × 3))}                                      \\
$128 \times h/2 \times w/2$ & \code{BatchNorm2D}                                      \\
$128 \times h/2 \times w/2$ & \code{LeakyReLU}                                        \\
$128 \times h/2 \times w/2$ & \code{UpSample (2×)}                                       \\
$64 \times h \times w$  & \code{SpectralNorm (Conv (3 × 3))}                                      \\
$64 \times h \times w$  & \code{BatchNorm2D}                                      \\
$64 \times h \times w$  & \code{LeakyReLU}                                        \\
$64 \times h \times w$     & \code{UpSample (2×)}                                     \\
$3 \times h \times w$      & \code{SpectralNorm (Conv (3 × 3))}                                      \\
$3 \times h \times w$      & \code{Sigmoid}                                          \\
$3 \times h \times w$      & \code{BatchNorm2D}                                      \\ \bottomrule
\end{tabular}
\end{adjustbox}
\caption{Architecture of the Generator Network ($\mathcal{G}$), detailing the sequence of operations and layer sizes from input to output. The network includes linear transformations, spectral normalization in convolution layers, batch normalization, leaky ReLU activations, upsampling, and a sigmoid activation for the output. Output dimensions at each layer are shown in relation to the input height (h) and width (w), with intermediate feature maps gradually upscaled to the final $3 \times h \times w$ generated image.}
\label{tab:g_arch}
\end{table}

\begin{table*}[t]
\renewcommand{\arraystretch}{1.5} 
\setlength{\tabcolsep}{8pt}      
\begin{adjustbox}{width=1\linewidth}
\begin{tabular}{|l|c|c|c|c|c|c|c|}
\hline
\multicolumn{1}{|c|}{}                 & Method & Image Resolution    & Synthetic Batch Size & $S$                       & $I$                   & $g$                    & Epoch $\mathcal{E}$                              \\ \hline
\multirow{2}{*}{ImageNettee/ImageWoof} & MUSE-S & $96 \times 96$            & 100                    & \multirow{2}{*}{50$\times d_r$}  & \multirow{2}{*}{5}  & \multirow{2}{*}{100} & \multirow{2}{*}{100}  \\ \cline{2-4}
                                       & MUSE   & {[}$96 \times 96$, $112 \times 112${]}  & {[}50, 40{]}            &                         &                     &                      &                                         \\ \hline
\multirow{2}{*}{ImageNet1k}              & MUSE-S & $112 \times 112$           & 200                    & \multirow{2}{*}{200$\times d_r$} & \multirow{2}{*}{20} & \multirow{2}{*}{100} & \multirow{2}{*}{400} \\ \cline{2-4}
                                       & MUSE   & {[}$112 \times 112$, $128 \times 128${]} & {[}200, 150{]}          &                         &                     &                      &                                          \\ \hline
\multirow{2}{*}{CIFAR10/CIFAR100}      & MUSE-S & $28 \times 28$            & 260                    & \multirow{2}{*}{2$\times d_r$}   & \multirow{2}{*}{20} & \multirow{2}{*}{40}  & \multirow{2}{*}{400} \\ \cline{2-4}
                                       & MUSE   & {[}$28 \times 28$, $32 \times 32${]}   & {[}130, 100{]}          &                         &                     &                      &                                          \\ \hline
\end{tabular}
\end{adjustbox}
\caption{The hyperparameters used in our methods across five different datasets are detailed below. \textbf{Image Resolution} and \textbf{Synthetic Batch Size} refer to the resolution and batch size of synthetic images generated by our methods. Notably, in the case of MUSE, two different resolutions are used, and their batch sizes are adjusted based on their scales. Other key parameters include: $S$, the number of training steps for optimizing the student model, scaled based on the data memory ratio ($d_r$); $I$, the number of times a batch of images is generated per epoch; and $g$, the training steps for optimizing the generators. Additionally, the following hyperparameters were fixed for all experiments: $\alpha_{ce} = 0.5$, $\alpha_{bn} = 10$, $\alpha_{adv} = 1.3$ (as in \cite{nayer}). Furthermore, in our paper, we propose the following parameters, which are also fixed for all experiments (their parameter sensitivity analysis can be found in Section \ref{sec:ps}): $\alpha_{cam} = 0.1$ (for CAM-Enhanced Quality Loss); $\alpha_{ed} = 10$, $\alpha_{aed} = 5$, $r_i = 0.015$, and $r_o = 0.03$ (for Embedding Diversity).}
\label{tab:hyperpara}
\end{table*}

\subsection{Student Model Training Details}

In all experiments, we adopt a consistent approach for training the student model. The batch size is set to match the Synthetic Batch Size, and the AdamW optimizer is used with a momentum of 0.9 and an initial learning rate of 1e-3. To further optimize training, a lambda scheduler is employed to adjust the learning rate dynamically throughout the training process.

\subsection{Other Settings}

We trained the model for $\mathcal{E}$ epochs, incorporating a warm-up phase during the first 10\% of $\mathcal{E}$, as outlined in the settings defined in \cite{fastdfkd, nayer}. This warm-up phase gradually increases the learning rate to stabilize training early on. Additionally, the model was trained with the specified batch size and other hyperparameters, which were carefully selected to ensure optimal performance. Further details regarding these parameters, including their values and any adjustments made during the training process, are provided in Table \ref{tab:hyperpara}.

\section{Parameter Sensitivity Analysis}
\label{sec:ps}

\noindent
\textbf{Parameter $\alpha_{cam}$.} In Table \ref{tab:pr-cam}, we compare the impact of different scale factors on CAM-Enhanced Quality Loss. The results show that our methods perform well, achieving higher accuracy with smaller scaling factors, peaking at a scale factor of 0.1. This can be attributed to the fact that the value of the CAM function is high due to direct subtract function, and a smaller scale factor is more effective for normalizing it.

\begin{table}[h]
\begin{adjustbox}{width=\linewidth}
\begin{tabular}{@{}lccccccc@{}}
\toprule
$\alpha_\text{cam}$             & 0.05  & 0.1            & 0.2   & 0.5   & 1     & 2     \\ \midrule
ImageNette (5\%)  & 79.77 & \textbf{80.32} & 80.2  & 79.69 & 78.26 & 78.63 \\
ImageNette (10\%) & 86.32 & \textbf{86.67} & 86.18 & 85.64 & 85.16 & 85.41  \\
ImageWoof (5\%)   & 36.03 & \textbf{36.25} & 35.67 & 35.66 & 35.13 & 35.11  \\
ImageWoof (10\%)  & 59.83 & \textbf{59.85} & 59.75 & 59.75 & 58.17 & 57.92 \\ \bottomrule
\end{tabular}
\end{adjustbox}
\caption{Comparison of the impact of various scale factors on CAM-Enhanced Quality Loss, highlighting the optimal performance achieved with smaller scale factors, peaking at a scale factor of 0.1.}
\label{tab:pr-cam}
\end{table}

\noindent
\textbf{Parameters $\alpha_{ed}$ and $\alpha_{aed}$.} Tables \ref{tab:pr-ed} and \ref{tab:pr-aed} compare the performance of different values of $\alpha_\text{ed}$ and $\alpha_\text{aed}$ on the ImageNette and ImageWoof datasets at 5\% and 10\% data memory ratio.In both tables, the highest accuracy is typically observed at intermediate values of $\alpha$, with $\alpha_\text{ed} = 10$ and $\alpha_\text{aed} = 5$ yielding the best results in most cases. his can be attributed to the fact that at these values, the mean squared error (MSE) distance between embeddings is significantly small. For instance, the minimum distance between two label text embeddings is just 0.03, which necessitates a higher scaling factor to amplify the impact of this term.

\begin{table}[]
\begin{adjustbox}{width=\linewidth}
\begin{tabular}{@{}lcccccc@{}}
\toprule
$\alpha_\text{ed}$             & 1     & 2     & 5     & 10             & 20             & 50      \\ \midrule
ImageNette (5\%)  & 80.02 & 79.62 & 79.59 & \textbf{80.12} & {79.79} & 80.27  \\
ImageNette (10\%) & 86.18 & 86.36 & 86.52 & \textbf{86.77} & {85.91} & 86.64 \\
ImageWoof (5\%)   & 35.77 & 35.69 & 36.12 & \textbf{36.31} & {35.37} & 36.13 \\
ImageWoof (10\%)  & 59.52 & 59.54 & 59.82 & \textbf{59.91} & {58.60}  & 59.70  \\ \bottomrule
\end{tabular}
\end{adjustbox}
\caption{Performance comparison of different $\alpha_\text{ed}$ values on the ImageNette and ImageWoof datasets at 5\% and 10\% sampling rates. The highest accuracy is achieved at $\alpha_\text{ed} = 10$, highlighting the importance of balancing the scaling factor to minimize MSE distance between embeddings.}
\label{tab:pr-ed}
\end{table}

\begin{table}[]
\begin{adjustbox}{width=\linewidth}
\begin{tabular}{@{}lccccccc@{}}
\toprule
$\alpha_\text{aed}$             & 1     & 2     & 5              & 10    & 20    & 50    \\ \midrule
ImageNette (5\%)  & 79.85 & 79.56 & \textbf{80.42} & 79.88 & 80.27 & 80.25 \\
ImageNette (10\%) & 86.58 & 86.01 & \textbf{86.68} & 85.56 & 86.17 & 85.67 \\
ImageWoof (5\%)   & 35.72 & 35.28 & \textbf{36.31} & 35.83 & 35.04 & 35.96 \\
ImageWoof (10\%)  & 59.35 & 58.88 & \textbf{59.88} & 59.54 & 59.47 & 59.52 \\ \bottomrule
\end{tabular}
\end{adjustbox}
\caption{Table \ref{tab:pr-aed}: Performance comparison of different $\alpha_\text{aed}$ values on the ImageNette and ImageWoof datasets at 5\% and 10\% sampling rates. Peak accuracy is observed at $\alpha_\text{aed} = 5$, emphasizing the role of scaling to optimize the MSE distance between embeddings.}
\label{tab:pr-aed}
\end{table}

\noindent
\textbf{Inner Radius $r_i$ and Outer Radius $r_o$.} In this approach, we follow the method proposed in \cite{lander} to determine the most effective radius. Based on this, we found that the minimum distance between two label text embeddings is 0.03. Therefore, we define the inner and outer radii around this value. As shown in Table \ref{tab:pr-r}, the pair of 0.015 ($r_i$) for the inner radius and 0.03 ($r_o$) for the outer radius yields the highest accuracy. This demonstrates that half of the minimum distance is optimal for the inner radius of Bounding Loss, similar to \cite{lander}, while the full minimum distance serves as the most effective outer radius. 

\begin{table}[]
\begin{adjustbox}{width=\linewidth}
\begin{tabular}{|c|c|c|c|c|c|}
\hline
\diagbox[width=\dimexpr \textwidth/12+2\tabcolsep\relax, height=0.6cm]{ $r_o$ }{$r_i$} & \textbf{0.05}  & \textbf{0.015}          & \textbf{0.03}  & \textbf{0.05}  & \textbf{0.1}   \\ \hline
\textbf{0.01 }& 76.30  & 80.21          & 80.20  & 80.13 & 76.33 \\ \hline
\textbf{0.03} & 77.33 & \textbf{80.44} & 80.24 & 79.08 & 76.32 \\ \hline
\textbf{0.1}  & 79.17 & 79.07          & 79.16 & 77.46 & 76.42 \\ \hline
\textbf{0.3}  & 78.09 & 78.19          & 78.02 & 77.41 & 76.36 \\ \hline
\textbf{1}    & 76.37 & 76.32          & 76.46 & 76.43 & 76.37 \\ \hline
\end{tabular}
\end{adjustbox}
\caption{Comparison of different inner ($r_i$) and outer ($r_o$) radius pairs for Bounding Loss and Marging Loss for Embedding Diversity terms. The pair of 0.015 for the inner radius and 0.03 for the outer radius achieves the highest accuracy, demonstrating that half of the minimum distance between embeddings works best for the inner radius, while the full minimum distance is optimal for the outer radius.}
\label{tab:pr-r}
\end{table}

\section{Extended Results}

\noindent
\textbf{Comparasion with Dataset Distillation.} To further demonstrate the benefits of our method, we compares on MUSE and MUSE-S against the SRe$^2$L \cite{sre2l} dataset distillation method on three datasets—Imagenetee, Imagewoof, and ImageNet1k, as shown in Table \ref{tab:dd} — using 10 and 50 images per class (IPC). MUSE achieves the highest accuracy in most settings; for instance, on Imagenetee with 10 IPC, MUSE reaches 35.2\% accuracy, surpassing (SRe$^2$L)’s 29.4\%. Similarly, on Imagewoof with 50 IPC, MUSE attains 37.3\% compared to (SRe$^2$L)’s 23.3\%. Although (SRe$^2$L) performs best on ImageNet1k with 50 IPC (46.8\%), MUSE generally shows strong performance, especially in lower IPC scenarios, underscoring its advantage in dataset distillation.

\begin{table}[h]
\centering
\begin{adjustbox}{width=\linewidth}
\begin{tabular}{@{}lcccccc@{}}
\toprule
Dataset  & \multicolumn{2}{c}{Imagenetee}            & \multicolumn{2}{c}{Imagewoof}             & \multicolumn{2}{c}{ImageNet1k}            \\ \midrule
IPC      & 10                  & 50                  & 10                  & 50                  & 10                  & 50                  \\
SReL     & 29.4 ± 3.0          & 40.9 ± 0.3          & 20.2 ± 0.2          & 23.3 ± 0.3          & 21.3 ± 0.6          & \textbf{46.8 ± 0.2} \\
MUSE-S     & {\ul 34.5  ± 1.3}   & {\ul 80.3 ± 0.2}    & {\ul 20.5 ± 0.2}    & {\ul 36.2 ± 0.4}    & {\ul 22.3 ± 0.7}    & 40.8 ± 0.3          \\
MUSE & \textbf{35.2 ± 1.6} & \textbf{82.2 ± 0.2} & \textbf{21.1 ± 0.2} & \textbf{37.3 ± 0.3} & \textbf{24.1 ± 0.5} & {\ul 42.8 ± 0.5}    \\ \bottomrule
\end{tabular}
\end{adjustbox}
\caption{Compared our methods with Dataset Distillation Methods (SRe$^2$L) \cite{sre2l} in same number of image per classes (IPC).}

\label{tab:dd}
\end{table}

\noindent
\textbf{Results on Semantic Segmentation.} To further assess the generalization capability of our method, we compare MUSE's performance with existing DFKD methods on the NYUv2 dataset. In this evaluation, we use a resolution of $128 \times 128$, while the other methods operate at $256 \times 256$. The results in Table \ref{tab:seg} show that our method consistently outperforms all other methods in semantic segmentation.

\begin{table}[h]

\centering
\begin{adjustbox}{width=\linewidth}
\begin{tabular}{@{}lccccc@{}}
\toprule
Method         & DFAD  & DAFL  & Fast  & NAYER & MUSE  \\ \midrule
Synthetic Time & 6.0h  & 3.99h & 0.82h & 0.82h & 0.82h \\
mIoU           & 0.364 & 0.105 & 3.66  & 3.85  & \textbf{4.01}  \\ \bottomrule
\end{tabular}
\end{adjustbox}

\caption{Performance comparison of MUSE with existing DFKD methods on the NYUv2 dataset.}

\label{tab:seg}
\end{table}

\section{Further Discussion.}
\noindent
\textbf{Choosing Target Mask  $M_{target}$.} In this section, we compared the performance of different target masks ($M_{\text{target}}$) across various sampling ratios (1\%, 5\%, 10\%, and 20\%). The target masks include Full($n$), where the matrix is filled with the value $n$, and G($i,j$), representing Gaussian matrices with a maximum value of $i$ and a standard deviation of $j$. As shown in Table \ref{tab:m}, the "Full(1)" matrix consistently outperforms other configurations, achieving the highest accuracy at all sampling ratios. While the "G(2,3)" and "G(1,3)" matrices exhibit similar performance, they are generally outperformed by "Full(1)" at most ratios. This indicates that filling the matrix with a constant value is the most effective approach for this task.

\begin{table}[h]
\centering
\begin{adjustbox}{width=\linewidth}
\begin{tabular}{@{}l|ccccccccc@{}}
\toprule
Ratio & Full(1)        & Full(2) & Full(3) & G(3,2) & G(3,3) & G(2,2) & G(2,3) & G(1,2) & G(1,3) \\ \midrule
1\%  & \textbf{34.52} & 34.4    & 33.11   & 34.32        & 34.26        & 33.3         & 33.3         & 34.49        & 33.94        \\
5\%  & \textbf{80.32} & 79.99   & 78.68   & 80.11        & 79.39        & 79.52        & 78.6         & 79.67        & 79.96        \\
10\%   & \textbf{86.67} & 86.53   & 86.24   & 86.31        & 86.44        & 85.7         & 86.56        & 86.12        & 85.88        \\
20\%   & \textbf{88.25} & 88.11   & 87.63   & 88.25        & 87.38        & 87.45        & 88.07        & 87.84        & 87.85        \\ \bottomrule
\end{tabular}
\end{adjustbox}
\caption{Performance Comparison Between Different Target Mask $M_{target}$. In that, Full(n) indidate matrix is fill by n and G(i,j) mean the Gaussian Matrix with max value of $i$ and $\sigma = j$}
\label{tab:m}
\end{table}

\noindent
\textbf{Choosing Class Representative Embedding $\vf_{\vy}$.}We evaluate the impact of using Label Text Embedding (LTE) and Class Center (CC) as the Class Representative Embedding $\vf_{\vy}$. The results in Table \ref{tab:cre} show that MUSE consistently outperforms NAYER across all settings. Furthermore, the performance of LTE and CC is comparable, with LTE exhibiting a slight advantage in some cases. This demonstrates the effectiveness of both configurations, providing flexibility in selecting between Class Center and Label Text Embedding representations.

\begin{table}[h!]
\centering
\begin{adjustbox}{width=\linewidth}
\begin{tabular}{@{}l|ccc|ccc@{}}
\toprule
Dataset                   & \multicolumn{3}{c|}{Imagenetee}       & \multicolumn{3}{c}{Imagewoof}        \\ 
Teacher                   & \multicolumn{3}{c|}{Resnet34 (94.06)} & \multicolumn{3}{c}{Resnet34 (83.02)} \\
Student                   & \multicolumn{3}{c|}{Resnet18 (93.53)} & \multicolumn{3}{c}{Resnet18 (82.59)} \\ \midrule
Ratio                     & 1\%        & 5\%        & 10\%       & 1\%         & 5\%        & 10\%      \\ 
NAYER                     & 9.35       & 32.17      & 42.57      & 6.72        & 15.62      & 25.27     \\
MUSE (CC) & 34.43      & 80.22      & 86.43      & 20.35       & 36.21      & 59.35     \\
MUSE (LTE)            & 34.51      & 80.36      & 86.61      & 20.47       & 36.41      & 59.62     \\ \bottomrule
\end{tabular}
\end{adjustbox}
\caption{Performance comparison of MUSE (using Class Center (CC) and Label Text Embedding (LTE)).}
\label{tab:cre}
\end{table}

\section{Lower-resolution Image for Vision Transformer.}

To illustrate the patch-reduction strategy mathematically, consider the input image resolution \( H \times W \). The Vision Transformer (ViT) splits the image into patches of size \( P \times P \), resulting in a grid of \( \frac{H}{P} \times \frac{W}{P} \) patches. For the standard ViT, with \( P = 16 \), and full-resolution images \( H = 224 \) and \( W = 224 \), the number of patches is:  

\begin{equation}
N_{\text{patches}} = \frac{H}{P} \cdot \frac{W}{P} = \frac{224}{16} \cdot \frac{224}{16} = 14 \cdot 14 = 196.
\end{equation}

For our approach, we reduce the resolution to \( H = 112 \) and \( W = 112 \), while maintaining \( P = 16 \). This results in:  

\begin{equation}
N_{\text{patches}} = \frac{H}{P} \cdot \frac{W}{P} = \frac{112}{16} \cdot \frac{112}{16} = 7 \cdot 7 = 49.
\end{equation}

\noindent
\textbf{Position Embedding.} Let the index matrix $ \mathcal{I} $ be a $ 10 \times 10 $ grid, where both row and column values range from 2 to 12:

\[
\mathcal{I} = \{(r, c) \mid 2 \leq r \leq 12, 2 \leq c \leq 12\}.
\]

We randomly select the center index $ p_\text{center} = ( p_\text{center}^r,  p_\text{center}^c) $ from this grid with a bias toward the center, particularly around indices $ 7 $ and $ 8 $ for both rows and columns. The probability of selecting the center index $ p_\text{center} $ is given by:

\[
P(p_\text{center}) \propto \frac{1}{1 + \lambda \cdot \left( | p_\text{center}^r - 7|^2 + | p_\text{center}^c - 7|^2 \right)},
\]

where:
\begin{itemize}
    \item $ ( p_\text{center}^r,  p_\text{center}^c) $ are the indices in the grid,
    \item $ \lambda $ is a parameter that controls the steepness of the decay, influencing how strongly the selection is biased toward the center,
    \item $ | p_\text{center}^r - 7|^2 + | p_\text{center}^c - 7|^2 $ represents the squared Euclidean distance from the center index $ (7, 7) $.
\end{itemize}

This formulation ensures that the selection probability decreases as the distance from the center increases, making the center indices $ (7, 7) $ and $ (8, 8) $ more likely to be chosen.

\noindent
\textbf{Patch Index Mapping.} After selecting the center index $ p_\text{center} = (r, c) $, the synthetic image patches are indexed relative to $ p_\text{center} $. Let $ p_i $ represent the index of the patch. The patch indices $ p_i $ are determined by an offset from $ p_\text{center} $. For a patch size of $ P \times P $, the patch index $ p_i $ is defined as:

\[
p_i = (p_\text{center}^r + \Delta r, p_\text{center}^c + \Delta c),
\]

where $ \Delta r, \Delta c \in \{-P, 0, P\} $ and are the offsets applied to the center index $  p_\text{center} $. This allows the selection of patches in a surrounding area around the center index $  p_\text{center} $.
This approach ensures 
that patch indices closer to the center are more likely to be selected, with the probability decreasing as the distance from the center increases. By prioritizing the center position embedding, our method, as demonstrated in Table \ref{tab:vit}, outperforms the original NAYER training, showing improvements of more than two percentage points.

\begin{table}[h!]
\centering
\begin{adjustbox}{width=\linewidth}
\begin{tabular}{@{}l|cc|cc@{}}
\toprule
Ratio  & \multicolumn{2}{c|}{1\%}           & \multicolumn{2}{c}{5\%}           \\ \midrule
Metric (Accuracy) & Top 1 (\%) & Top 5 (\%) & Top 1 (\%) & Top 5 (\%) \\ 
NAYER  & 4.52           & 19.45          & 16.24          & 43.24          \\
MUSE   & 15.24          & 36.52          & 28.24          & 60.24          \\ \bottomrule
\end{tabular}
\end{adjustbox}
\caption{Performance Comparison Our MUSE and NAYER in ViT model}
\label{tab:vit}
\end{table}
\end{document}